# Supervised Dictionary Learning by a Variational Bayesian Group Sparse Nonnegative Matrix Factorization

Ivan Ivek

**Abstract**— Nonnegative matrix factorization (NMF) with group sparsity constraints is formulated as a probabilistic graphical model and, assuming some observed data have been generated by the model, a feasible variational Bayesian algorithm is derived for learning model parameters. When used in a supervised learning scenario, NMF is most often utilized as an unsupervised feature extractor followed by classification in the obtained feature subspace. Having mapped the class labels to a more general concept of groups which underlie sparsity of the coefficients, what the proposed group sparse NMF model allows is incorporating class label information to find low dimensional label-driven dictionaries which not only aim to represent the data faithfully, but are also suitable for class discrimination. Experiments performed in face recognition and facial expression recognition domains point to advantages of classification in such label-driven feature subspaces over classification in feature subspaces obtained in an unsupervised manner.

## 1 Introduction

SINCE the appearance of the seminal paper [1], NMF has become a popular data decomposition technique due to succesful applications in a still growing number of fields where data are nonnegative, such as pixel intensities in computer vision, amplitude spectra in audio signal analysis and EEG signal analysis, term counts in document clustering problems, and item ratings in collaborative filtering.

NMF aims at decompositions $\boldsymbol{X} \approx \boldsymbol{TV}$, where $\boldsymbol{X}$, $\boldsymbol{T}$ and $\boldsymbol{V}$ are all nonnegative matrices. Throughout this paper $\boldsymbol{X}$ will be regarded as a collection of data samples organized columnwise, $\boldsymbol{T}$ as a dictionary of features organized columnwise, and $\boldsymbol{V}$ as matrix of coefficients when $\boldsymbol{X}$ is projected onto the dictionary $\boldsymbol{T}$. Under assumptions of linearity and nonnegativity, when underlying dimensionality is lower than dimensionality of the original space of the data $\boldsymbol{X}$, dimensionality reduction of the data can effectively be achieved this way.

Although the decomposition is nonunique in general, NMF is able to produce strictly additive decompositions perceived as part-based by adding additional bias in the model [1], [2]. To this end, different sparsity promoting regularizers have been proposed for divergence-based NMF [3]. Also, to include higher order data descriptions, many other variants have been developed, e.g. Local NMF [4] with locality constraints, Non-smooth NMF [5] with regularization for sparse and localized features, NMF with smoothness constraints [6], graph regularized NMF [7], [8] manifold regularized NMF [9]. More recently, alternative formulations of NMF in probabilistic framework have been developed [10], [11], allowing for explicit modeling of richer structural constraints as graphical models [12], [13], [14], [15].

### 1.1 Context and Contributions

Closely related to work in this paper are NMF formulations and extensions with group sparsity constraints. Addressing EEG classification where problem is to classify different tasks being performed by different subjects, in [16] a divergence based NMF with mixed-norm regularization imposed on dictionary elements to to get task-related (common) features that are as close as possible and, separatedly, features which reflect task-independent (individual) characteristics, required to be as far as possible, has been proposed. Another divergence based algorithm but with group sparsity penalizations on the coefficient matrix has been proposed in [17]. In [18] a generative model with the same aim, enforcing two groups of features with the previously mentioned properties, is trained using a variational Bayesian approach. In [19] a NMF variant with Itakura-Saito divergence with a direct group-sparsity enforcing penalization of coefficient matrix has been succesfully applied to blind audio source separation. Very recently, in [20] a generative model trained with Markov chain Monte Carlo is proposed to separate features into groups of common bases and individual bases with Laplacian scale mixture distributions as priors for the two groups, and applied to blind music source separation.

Work presented in this paper had come out of the probabilistic NMF modeling track of research, with group sparsity constraints as exponential scale mixture distributions imposed on coefficient matrix directly rather than seeking group sparsity by constraining the two groups of features as in [16], [18], [20]. Comparatively, it may be

• *Ivek, I. is with the Division of Electronics, Rudjer Boskovic Institute, Croatia. E-mail: ivan.ivek@irb.hr*

noted that algorithms [16], [17], [19] are not of Bayesian type. Although it aims to impose group sparsity through common and individual features, most resemblance to the here presented model bears [20] because of Laplacian scale mixture as a prior for the two groups, but it uses Metropolis-Hastings algorithm for parameter estimation due to choices of distributions in their hierarchical model. Group sparse NMF presented in this paper however, is engineered in a way that it allows efficient learning, here performed using mean-field variational Bayesian methodology, which is deterministic and features explicit variational bound calculation based upon which model comparison and selection can be made [11].

Most commonly, NMF is utilized in two scenarios: unsupervised learning, to get decompositions suitable for clustering, and supervised learning, where NMF is used as an unsupervised feature extractor followed by classification in the obtained feature space. In both those cases NMF does not incorporate label information. Applications presented in this paper are focused on the latter case, but, instead of ignoring the data labeling, the proposed group sparse NMF model is put in a setup where it is label-driven in an attempt to bring out potential information in the labeling to find feature subspaces which aid classification. One remaining scenario is NMF in semi-supervised learning applications, which divergence based algorithms [21], [22], [23], [24] have been developed for, also possessing the ability to include and use label information.

### 1.2 Probabilistic Formulations of NMF

It has been shown in [10] that maximum-likelihood (ML) estimation of factor matrices $\boldsymbol{T}$ and $\boldsymbol{V}$ under different noise distributions are equivalent to NMF algorithms with different corresponding divergences. More precisely, ML estimation of probabilistic NMF under Gaussian, Poissonian, and Gamma noise correspond to minimization of Euclidean, Kullback-Leibler (KL) and Itakura-Saito divergences, respectively. In case of maximum-a-posteriori (MAP) estimation, exponential prior on a factor corresponds to sparsity promoting l1 regularization. Another connection between probabilistic modeling and NMF worth noting is that Probabilistic Latent Semantic Analysis, which is an expectation-maximization algorithm, is equivalent to KL-NMF algorithm using multiplicative updates [25]. Apart from ML, MAP and EM estimators, Bayesian methods (Monte Carlo, variational Bayes) have successfully been employed for efficient NMF parameter learning [11], [14], [15].

### 1.3 Variational Bayesian Learning

Consider the objective of minimizing dissimilarity between conditional distribution $P(H|D,\Theta)$, where $H$ denotes hidden variables in the model, $D$ observed variables and $\Theta$ model parameters, and its instrumental variational approximation $q(H)$, quantified by Kullback-Leibler divergence

$$\mathrm{KL}(q||p) = \int_H q(H)\log\left(\frac{q(H)}{p(H|D,\Theta)}\right)\mathrm{d}H. \tag{1}$$

Equation (1) can be be rewritten as

$$\begin{aligned}\mathrm{KL}(q||p) &= \int_H q(H)\log\left(\frac{q(H)}{p(H|D,\Theta)}\right)\mathrm{d}H\\ &= \int_H q(H)\log\left(\frac{q(H)p(D|\Theta)}{p(H,D|\Theta)}\right)\mathrm{d}H\\ &= \int_H q(H)\log\left(\frac{q(H)}{p(H,D|\Theta)}\right)\mathrm{d}H + \int_H q(H)p(D|\Theta)\,\mathrm{d}H\\ &= \int_H q(H)\log\left(\frac{q(H)}{p(H,D)}\right)\mathrm{d}H + \log p(D|\Theta)\\ &= -\mathcal{L}(q) + \log p(D|\Theta).\end{aligned}$$

Because KL divergence is nonnegative, it turns out that $\mathcal{L}(q)$ is lower bound on the marginal loglikelihood of the observed data. On the other hand, because $\log p(D|\Theta)$ is constant, the objective of minimizing KL divergence can be reformulated as maximization of $\mathcal{L}(q)$.

By expanding the expression for the variational bound

$$\begin{aligned}\mathcal{L}(q) &= \int_H q(H)\log p(H,D)\,\mathrm{d}H - \int_H q(H)\log q(H)\,\mathrm{d}H\\ &= \langle\log p(H,D|\Theta)\rangle_{q(H)} + \mathcal{H}(q),\end{aligned} \tag{2}$$

where $\mathcal{H}(q)$ denotes entropy of the approximation $q(H)$, and supposing that the approximative variational distribution takes a factorized form $q(H) = \prod_{\alpha\epsilon \mathrm{C}} q_\alpha$, it can be shown that iterative local updates alternating over C of form

$$(q_\alpha)^{(n+1)} \propto \exp\left(\langle\log p(H,D|\Theta)\rangle_{q_{\neg\alpha}{}^{(n)}}\right), \tag{3}$$

improve lower bound on the marginal loglikelihood monotonically, with

$$q_{\neg\alpha} = \frac{\prod_{\alpha'} q_{\alpha'}}{q_\alpha}.$$

Should the mode be conjugate-exponential, all the expectations in (3) necessarily assume analytical forms [26].

## 2 Methodology

### 2.1 Exponential Scale Mixture Distribution

Let random variable $V$ be a product of reciprocal of some positive random variable $\Lambda$ and exponentially distributed random variable $U$ with scale 1,

$$V = \Lambda^{-1}U.$$

Supposing independency of $\lambda$ and $u$, conditioned on $\lambda$, $v$ is exponentially distributed with scale $\lambda^{-1}$,

$$p(v|\lambda) = Exponential(v|\lambda^{-1}). \tag{4}$$

and its marginal distribution assumes form of a continuous mixture with mixing variable $\lambda$,

$$p(v) = \int_0^\infty p(v|\lambda)p(\lambda)\,\mathrm{d}\lambda.$$

Note that, should the distribution of the mixing variable be discrete, the above expression collapses to a discrete mixture of exponentials with specific shape parameters.

Because exponential distribution can be obtained by truncation only of Laplacian distribution, it follows that exponential scale mixture is a special case of Laplacian scale mixture distribution [20], [27].

### 2.2 Group Sparsity Model

Placing exponential scale mixture in setting of probabilistic inference and learning, feasibility of related algorithms depends on the choice of prior $p(\lambda)$ and, as feasibility is prioritized, this choice of suitable priors gets narrowed down. As (3) suggests, conjugacy is desireable when computational complexity is taken into consideration. For examples of how conjugate-exponential models are used for variational Bayesian NMF, the interested reader is referred to [11], [14] and also [15], where a variational family more general than one which conjugate model would suggest to get analytical expressions appropriate for optimization.

In this paper prior $p(\lambda)$ is engineered as a hierarchical graphical model with a discrete mixture

$$p(\lambda|\lambda_{\mathrm{c}}, z) = \delta\left(\lambda - \sum_{c=1}^{C} \delta(z-c)\lambda_{\mathrm{c}}\right) \quad (5)$$

of $C$ gamma distributed variables $\lambda_{\mathrm{c}}$ ,

$$p\left(\lambda_{\mathrm{c}}|a_{\mathrm{c}}^{\lambda}, b_{\mathrm{c}}^{\lambda}\right) = Gamma\left(\lambda_{\mathrm{c}}|a_{\mathrm{c}}^{\lambda}, b_{\mathrm{c}}^{\lambda}\right), \mathrm{c}\epsilon\{1,\dots,C\} \quad (6)$$

with $z$ as a categorical mixture selector variable.

According to (4), (5) and (6), $v$ is exponentially distributed with different scale parameters ${\lambda_{\mathrm{c}}}^{-1}$ for different selections of $z$.

To clarify on relationship to group sparsity, suppose first that multiple variables $v_\tau$ exist organized in groups $\{1,\dots,C\}$, their affiliation indicated by variables $z_\tau\epsilon\{1,\dots,C\}$. Because reciprocals of the variables $\lambda_c$, representing mean values of the exponential distributions in the mixture and being interpreted as continuous indicators of how large the averaged outcomes in a group are, have inverse gamma probability density functions, the masses of these density functions can be tuned to be concentrated on small values. Such priors act as constraints in a way that only the minority of the groups are expected to have exponential distributions with significantly large mean values, which is a suitable way to describe group sparse processes consistently in a probabilistic setup.

### 2.3 Probabilistic NMF with Group Sparsity Prior

The proposed generative NMF model consist of:

1) mixing stage with dimensionality reduction effect under Poissonian noise

$$p(x_{\mathrm{v\tau}}|s_{\mathrm{v:\tau}}) = \delta(x_{\mathrm{v\tau}} - \textstyle\sum_i s_{\mathrm{vi\tau}})$$

$$p(s_{\mathrm{vi\tau}}|t_{\mathrm{vi}}, v_{\mathrm{i\tau}}) = Poisson(s_{vi\tau}; t_{vi}v_{i\tau}) \ ,$$

2) gamma priors for the left matrix T

$$p(t_{vi}|a_{vi}^t, a_{vi}^t) = Gamma(t_{vi}|a_{vi}^t, b_{vi}^t)$$

3) group sparsity structural constraints over the right matrix $\boldsymbol{V}$

$$p(v_{\mathrm{i\tau}}|z_\tau, \lambda_{i:}) = Exponential(v_{\mathrm{i\tau}}|(\textstyle\sum_{\mathrm{c}} \delta(z_\tau - c)\lambda_{\mathrm{ic}})^{-1})$$

$$p\left(\lambda_{\mathrm{ic}}|a_{\mathrm{ic}}^{\lambda}, b_{\mathrm{ic}}^{\lambda}\right) = Gamma\left(\lambda_{\mathrm{ic}}|a_{\mathrm{ic}}^{\lambda}, b_{\mathrm{ic}}^{\lambda}\right).$$

Both the mixing stage and the gamma priors over $\boldsymbol{T}$ are designed as in [11], while the structural constraints over $\boldsymbol{V}$ are introduced as a novelty.

Joint distribution of the proposed model is

$$p(H, D|\Theta) = p\left(\boldsymbol{X}, \boldsymbol{S}, \boldsymbol{T}, \boldsymbol{V}, \boldsymbol{\Lambda}\middle|\boldsymbol{A^t}, \boldsymbol{B^t}, \boldsymbol{A^\lambda}, \boldsymbol{B^\lambda}, \vec{z}\right)$$

$$= p(\boldsymbol{X}|\boldsymbol{S})p(\boldsymbol{S}|\boldsymbol{T}, \boldsymbol{V})p(\boldsymbol{T}|\boldsymbol{A^t}, \boldsymbol{B^t})$$

$$* \, p\left(\boldsymbol{V}\middle|\boldsymbol{\Lambda}, \vec{z}\right)p(\boldsymbol{\Lambda}|\boldsymbol{A^\lambda}, \boldsymbol{B^\lambda}),$$

with

$$p(\boldsymbol{X}|\boldsymbol{S}) = \textstyle\prod_{v,\tau} p\,(x_{v\tau}|s_{v:\tau})$$

$$p(\boldsymbol{S}|\boldsymbol{T}, \boldsymbol{V}) = \textstyle\prod_{v,\tau} p\,(s_{v:\tau}|t_{vi}, v_{i\tau})$$

$$p(\boldsymbol{T}|\boldsymbol{A^t}, \boldsymbol{B^t}) = \textstyle\prod_{v,i} p\,(t_{vi}|a_{vi}^t, b_{vi}^t)$$

$$p\left(\boldsymbol{V}\middle|\boldsymbol{\Lambda}, \vec{z}\right) = \textstyle\prod_{\mathrm{i},\tau} p\,(v_{i\tau}|\lambda_{i:}, z_\tau)$$

$$p(\boldsymbol{\Lambda}|\boldsymbol{A^\lambda}, \boldsymbol{B^\lambda}) = \textstyle\prod_{\mathrm{i},c} p\left(\lambda_{\mathrm{ic}}|a_{\mathrm{ic}}^{\lambda}, b_{\mathrm{ic}}^{\lambda}\right).$$

### 2.4 Variational Learning Algorithm

In order to obtain convenient analytical forms of iterative updates, the variational distribution is chosen to be factorized as

$$q(H) = q(\boldsymbol{S}, \boldsymbol{T}, \boldsymbol{V}, \boldsymbol{\Lambda}) = q(\boldsymbol{S})q(\boldsymbol{T})q(\boldsymbol{V})q(\boldsymbol{\Lambda}),$$

with

$$q(\boldsymbol{S}) = \textstyle\prod_{v,\tau} q(s_{\mathrm{v:\tau}})$$

$$q(\boldsymbol{T}) = \textstyle\prod_{v,i} q(t_{\mathrm{vi}})$$

$$q(\boldsymbol{V}) = \textstyle\prod_{i,\tau} q(v_{\mathrm{i\tau}})$$

$$q(\boldsymbol{\Lambda}) = \textstyle\prod_{\mathrm{i},c} q(\lambda_{\mathrm{ic}}) \ .$$

As the proposed learning algorithm will be layed out in matrix form with computationally efficient matrix operations, hyperparameters and variational parameters of the model are organized as matrices according to Table 1 and Table 2, respectively.

| | |
|---|---|
| $[\boldsymbol{X}]_{\mathrm{v\tau}} = x_{\mathrm{v\tau}}$ | $[\boldsymbol{\Delta}]_{\tau c} = \delta(z_\tau - c)$ |
| $[\boldsymbol{A^t}]_{vi} = a_{vi}^t$<br>$[\boldsymbol{B^t}]_{vi} = b_{vi}^t$ | $[\boldsymbol{A^\lambda}]_{ic} = a_{\mathrm{ic}}^{\lambda}$<br>$[\boldsymbol{B^\lambda}]_{ic} = b_{\mathrm{ic}}^{\lambda}$ |

Table 1. Parameters and hyperparameters of the proposed model

| | |
|---|---|
| $\left[\boldsymbol{E_t^{(n)}}\right]_{vi} = \langle t_{vi}\rangle^{(n)}$<br>$\left[\boldsymbol{L_t^{(n)}}\right]_{vi} = \langle \log t_{vi}\rangle^{(n)}$ | $\left[\boldsymbol{E_v^{(n)}}\right]_{\mathrm{i\tau}} = \langle v_{\mathrm{i\tau}}\rangle^{(n)}$<br>$\left[\boldsymbol{L_v^{(n)}}\right]_{\mathrm{i\tau}} = \langle \log v_{\mathrm{i\tau}}\rangle^{(n)}$ |
| $\left[\boldsymbol{\Sigma_t^{(n)}}\right]_{vi} = \sum_\tau \langle s_{vi\tau}\rangle^{(n)}$<br>$\left[\boldsymbol{\Sigma_v^{(n)}}\right]_{i\tau} = \sum_v \langle s_{vi\tau}\rangle^{(n)}$ | $\left[\boldsymbol{E_\lambda^{(n)}}\right]_{ic} = \langle \lambda_{\mathrm{ic}}\rangle^{(n)}$<br>$\left[\boldsymbol{L_\lambda^{(n)}}\right]_{ic} = \langle \log \lambda_{\mathrm{ic}}\rangle^{(n)}$ |

Table 2. Variational parameters of the proposed model

To derive iterative alternating updates for $q(s_{\mathrm{v:\tau}})^{(n+1)}$, applying (5) yields

$$q(s_{\mathrm{v:\tau}})^{(n+1)} \propto \exp\langle \log p(H, D|\Theta)\rangle_{\frac{q(H)^{(n)}}{q(s_{v:\tau})^{(n)}}}$$

$$= Multinomial(s_{v:\tau}|x_{\mathrm{v\tau}}, {p_{\mathrm{vi\tau}}}^{(n)})$$

with natural parameters

$${p_{\mathrm{vi\tau}}}^{(n)} = \frac{\exp(\langle \log t_{vi}\rangle^{(n)} + \langle \log v_{\mathrm{i\tau}}\rangle^{(n)})}{\sum_i \exp(\langle \log t_{vi}\rangle^{(n)} + \langle \log v_{\mathrm{i\tau}}\rangle^{(n)})} \ .$$

Analytically, variational factor $q(s_{\mathrm{v:\tau}})^{(n+1)}$ assumed form of a multinomial distribution. Using analytical form of expectation of sufficient statistics of a multinomial distribution, they get updated according to

$$\langle s_{\mathrm{vi\tau}}\rangle^{(n+1)} = x_{\mathrm{v\tau}}\, {p_{\mathrm{vi\tau}}}^{(n)}. \quad (7)$$

Specifically, as will be seen later, alternating updates for other hidden variables in the model will require those expectations in forms $\sum_{v}\langle s_{vi\tau}\rangle^{(n+1)}$ and $\sum_{\tau}\langle s_{vi\tau}\rangle^{(n+1)}$, which, by putting summation over (7) and by simple algebraic manipulation, compactly become

$$\sum_{v}\langle s_{vi\tau}\rangle^{(n+1)} = \exp\langle\log v_{i\tau}\rangle^{(n)} \sum_{v} \exp\langle\log t_{vi}\rangle^{(n)} \xi_{v\tau}{}^{(n)} \quad (8)$$

$$\sum_{\tau}\langle s_{vi\tau}\rangle^{(n+1)} = \exp\langle\log t_{vi}\rangle^{(n)} \sum_{\tau} \xi_{v\tau}{}^{(n)} \exp\langle\log v_{i\tau}\rangle^{(n)}, \quad (9)$$

with repeating term substituted as

$$\xi_{v\tau}{}^{(n)} = \frac{x_{v\tau}}{\sum_{i} \exp\langle\log t_{vi}\rangle^{(n)} \exp\langle\log v_{i\tau}\rangle^{(n)}}. \quad (10)$$

Rewritten in matrix notation, expressions (8), (9) and (10) now become

$$\boldsymbol{\Sigma}_{v}^{(n+1)} = \exp \boldsymbol{L}_{v}^{(n)} .* (\exp \boldsymbol{L}_{t}^{(n)})^{T} * \boldsymbol{\xi}^{(n)})$$

$$\boldsymbol{\Sigma}_{t}^{(n+1)} = \exp \boldsymbol{L}_{t}^{(n)} .* (\boldsymbol{\xi}^{(n)} * (\exp \boldsymbol{L}_{v}^{(n)})^{T})$$

$$\boldsymbol{\xi}^{(n+1)} = \boldsymbol{X}./((\exp \boldsymbol{L}_{t}^{(n)}) * (\exp \boldsymbol{L}_{v}^{(n)})).$$

Approached in the same manner, $q(t_{vi})^{(n+1)}$ analytically assumes form of a gamma probability density function:

$$q(t_{vi})^{(n+1)} \propto \exp\langle\log p(H, D|\Theta)\rangle_{\frac{q(H)^{(n)}}{q(t_{vi})^{(n)}}} = Gamma\left(t_{vi}|\alpha_{vi}^{t\,(n)}, \beta_{vi}^{t\,(n)}\right)$$

with shape and scale parameters

$$\alpha_{vi}^{t\,(n)} = a_{vi}^{t} + \sum_{\tau}\langle s_{vi\tau}\rangle^{(n)}$$

$$\beta_{vi}^{t\,(n)} = \left((b_{vi}^{t})^{-1} + \sum_{\tau}\langle v_{i\tau}\rangle^{(n)}\right)^{-1},$$

respectively. Again, it suffices to update and store the expectations of sufficient statistics of $q(t_{vi})^{(n+1)}$,

$$\langle t_{vi}\rangle^{(n+1)} = \alpha_{vi}^{t\,(n)} \beta_{vi}^{t\,(n)}$$

$$\langle\log t_{vi}\rangle^{(n+1)} = \Psi(\alpha_{vi}^{t\,(n)}) + \log \beta_{vi}^{t\,(n)}.$$

In matrix notation, these updates assume forms

$$\boldsymbol{A_t}^{(n)} = \boldsymbol{1}.* \boldsymbol{A^t} + \boldsymbol{\Sigma}_{t}^{(n)}$$

$$\boldsymbol{B_t}^{(n)} = 1./(1./\boldsymbol{B^t} + \boldsymbol{1} * \boldsymbol{E}_{v}^{(n)^T})$$

$$\boldsymbol{E}_{t}^{(n+1)} = \boldsymbol{A_t}^{(n)} .* \boldsymbol{B_t}^{(n)}$$

$$\boldsymbol{L}_{t}^{(n+1)} = \Psi(\boldsymbol{A_t}^{(n)}) + \log \boldsymbol{B_t}^{(n)},$$

where by $\boldsymbol{1}$ a unity matrix of appropriate size is denoted and $\Psi(.)$ is elementwise digamma function.

Likewise, iterative update equations for $q(v_{i\tau})^{(n+1)}$ are derived as follows:

$$q(v_{i\tau})^{(n+1)} \propto \exp\langle\log p(H, D|\Theta)\rangle_{\frac{q(H)^{(n)}}{q(v_{i\tau})^{(n)}}} = Gamma\left(v_{i\tau}|\alpha_{i\tau}^{v\,(n)}, \beta_{i\tau}^{v\,(n)}\right),$$

$$\alpha_{i\tau}^{v\,(n)} = 1 + \sum_{v}\langle s_{vi\tau}\rangle^{(n)}$$

$$\beta_{i\tau}^{v\,(n)} = \left(\sum_{c} \delta(z_\tau - c)^{(n)} \langle\lambda_{ic}\rangle^{(n)} + \sum_{v}\langle t_{vi}\rangle^{(n)}\right)^{-1},$$

and the corresponding updates are to be done as

$$\langle v_{i\tau}\rangle^{(n+1)} = \alpha_{i\tau}^{v\,(n)} \beta_{i\tau}^{v\,(n)}$$

$$\langle\log v_{i\tau}\rangle^{(n+1)} = \Psi(\alpha_{i\tau}^{v\,(n)}) + \log \beta_{i\tau}^{v\,(n)},$$

or, in matrix form,

$$\boldsymbol{A_v}^{(n)} = \boldsymbol{1} + \boldsymbol{\Sigma}_{v}^{(n)}$$

$$\boldsymbol{B_v}^{(n)} = \boldsymbol{1}./(\boldsymbol{E}_{\lambda}^{(n)} * \boldsymbol{\Delta}^{(n)^T} + \boldsymbol{E}_{t}^{(n)^T} * \boldsymbol{1})$$

$$\boldsymbol{E}_{v}^{(n+1)} = \boldsymbol{A_v}^{(n)} .* \boldsymbol{B_v}^{(n)}$$

$$\boldsymbol{L}_{v}^{(n+1)} = \boldsymbol{\Psi}(\boldsymbol{A_v}^{(n)}) + \log \boldsymbol{B_v}^{(n)}.$$

Iterative update equations for $\Lambda$ are derived as follows:

$$q(\lambda_{ic})^{(n+1)} \propto \exp\langle\log p(H, D|\Theta)\rangle_{\frac{q(H)^{(n)}}{q(\lambda_{ic})^{(n)}}} \propto \sum_{\tau}\langle\log p(v_{i\tau}|\lambda_{i:}, z_\tau)\rangle + \langle\log p\left(\lambda_{ic}|a_{ic}^{\lambda}, b_{ic}^{\lambda}\right)\rangle,$$

where all the expectations are taken with respect to $q(v_{i\tau})^{(n)} \frac{\prod_{c} q(\lambda_{ic})^{(n)}}{q(\lambda_{ic})^{(n)}} q\left(\vec{z}\right)^{(n)}$, not explicitly noted for clarity.

The first term,

$$\sum_{\tau}\langle\log p(v_{i\tau}|\lambda_{i:}, z_\tau)\rangle \propto \sum_{\tau}(-\langle v_{i\tau}\rangle \sum_{c} \delta(z_\tau - c)\lambda_{ic}) + \sum_{\tau}\langle\log\ (\sum_{c} \delta(z_\tau - c)\lambda_{ic})\rangle,$$

is seemingly more difficult than what has been encountered so far due to expectation operator over the logarithmic function. Luckily, by observing that logarithmic function is concave, the lower bound can be relaxed using Jensen's inequality,

$$\langle\log(\sum_{c} \delta(z_\tau - c)\lambda_{ic}\rangle \geq \sum_{c} \delta(z_\tau - c)\langle\log \lambda_{ic}\rangle.$$

In this relaxed lower bound, coordinate ascent now admits a closed form

$$q(\lambda_{ic})^{(n+1)} = Gamma(\lambda_{ic}|\alpha_{ic}^{\lambda\,(n)}, \beta_{ic}^{\lambda\,(n)}),$$

$$\alpha_{ic}^{\lambda\,(n)} = a_{ic}^{\lambda} + \sum_{\tau} \delta(z_\tau - c)$$

$$\beta_{ic}^{\lambda\,(n)} = \left(b_{ic}^{\lambda^{-1}} + \sum_{\tau}\langle v_{i\tau}\rangle \delta(z_\tau - c)\right)^{-1}.$$

The corresponding expectations of natural parameters are updated according to

$$\langle\lambda_{ic}\rangle^{(n+1)} = \alpha_{ic}^{\lambda\,(n)} \beta_{ic}^{\lambda\,(n)}$$

$$\langle\log \lambda_{ic}\rangle^{(n+1)} = \Psi\left(\alpha_{ic}^{\lambda\,(n)}\right) + \log \beta_{ic}^{\lambda\,(n)},$$

which can compactly be rewritten in matrix form as

$$\boldsymbol{A_\lambda}^{(n)} = \boldsymbol{A}^{\lambda} + \boldsymbol{\Delta}^{(n)} * \boldsymbol{1}$$

$$\boldsymbol{B_\lambda}^{(n)} = 1./\left(1./\boldsymbol{B}^{\lambda} + \boldsymbol{E}_{v}^{(n)} * \boldsymbol{\Delta}^{(n)}\right)$$

$$\boldsymbol{E}_{\lambda}^{(n+1)} = \boldsymbol{A_\lambda}^{(n)} .* \boldsymbol{B_\lambda}^{(n)}$$

$$\boldsymbol{L}_{\lambda}^{(n+1)} = \Psi(\boldsymbol{A_\lambda}^{(n)}) + \log \boldsymbol{B_\lambda}^{(n)}.$$

The learning algorithm in matrix form is recapitulated in Appendix C of the paper, with the iterative updates in the same order as presented above.

In the presented algorithm, group affiliation variables $z_\tau$ are assumed to be fully observed, i.e. groups are to be explicitly defined beforehand depending on the application, as the model does not attempt to learn the group affiliations.

## 2.5 Variational Bound

Expression for the variational lower bound on the marginal loglikelihood of the observed data is obtained by expanding (2), as presented in Appendix B.

## 3 EXPERIMENTS

In this subsection performance of a simple classification method based on the proposed group sparse NMF algorithm is investigated on three publicly available benchmark datasets, Yale [28] and ORL [29], [30] for face recognition and JAFFE [31] for facial expression recognition. Yale dataset consists of 11 grayscale images per subject of 15 subjects, one image per different facial expression of configuration: center-light, w/glasses, happy, left-light, w/no glasses, normal, right-light, sad, sleepy, surprised and wink. ORL dataset consists of 10 different images of 40 distinct subjects, taken at different times, varying the lightning, facial expressions (open or close eyes, smiling or not smiling) and facial details (glasses or no glasses). JAFFE dataset is a collection of 213 images of 7 facial expressions - angry, disgust, fear, happy, neutral, sad surprise - posed by 10 Japanese female models.

Results are compared to classification methods based on PCA and related NMF algorithms, and also, where available, to relevant published results on the same datasets with different approaches.

MATLAB/Octave implementation of the algorithm as well as scripts used to generate the results are available from the author's homepage or will have been received upon request.

### 3.1 Data Preprocessing

Images in Yale dataset used in the experiments have been prepared by the MIT media laboratory [32] - aligned by rotation and centering of the manually determined locations of the eyes and then cropped. Additionaly, specifically for this paper, after downsampling the images by factor 0.5 to alleviate computational load, pixelwise masking has been applied to remove most of the background, torso and the hair. Finally, histograms of masked images have been equalized.

In case of ORL dataset, because faces have been taken at different angles, centering has not been attempted. Images have only been downsampled by factor of 0.5, with histogram equalization.

Images from JAFFE dataset have first been roughly aligned by congealing [33], [34], then resized by factor 0.75 and masked leaving only pixels which roughly correspond to locations of faces, followed by histogram equalization.

For all datasets, images have been vectorized in a way that each image is a column vector of input matrix which is to be factorized. Examples of preprocessed images are shown in Fig. 1.

### 3.2 Experimental Setting

The classification method consists of three consecutive stages: image preprocessing stage, dictionary learning stage and classification stage. Preprocessing is dependent on the dataset and has been described in detail in the preceeding subsection. In the dictionary learning stage a dictionary is obtained by the proposed group sparse NMF algorithm, or PCA or the standard sparse NMF algorithms; only in case of the probabilistic group sparsity NMF algorithm are the class labels taken into account, while in other cases the algorithms cannot include this information straightforwardly and is therefore done in a fully unsupervised manner, which is exactly where the comparative advantage of the proposed algorithm for classification problems lies.

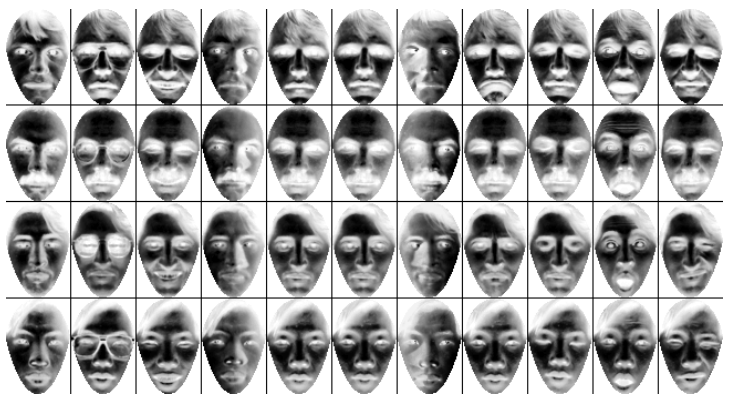

a)

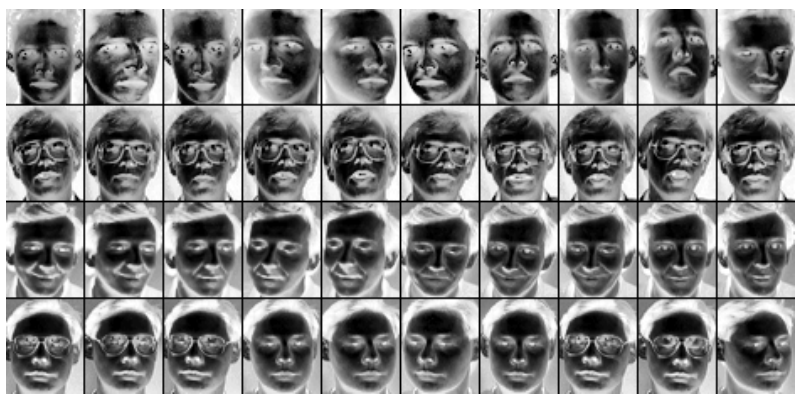

b)

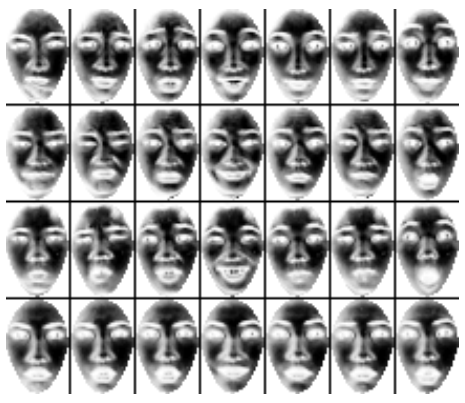

c)

Fig. 1. Examples of preprocessed face samples (in negative): a) Yale, b) ORL, c) JAFFE.

To put more weight on role of the quality of the decomposition obtained in the dictionary learning stage, in a sense of class-to-class discriminative information it is able to bring out by itself rather than on the role of the classifier, the simple classifier of choice in the classification stage is 1-nearest neighbor with cosine distance metric.

Performance quantifiers have been obtained by 5 runs of 10-fold crossvalidation for 10 different random restarts of NMF algorithms; at each pass dictionary learning by one of the algorithms had been performed on the training set only, followed by obtaining the representation of the test set in the feature space (in which the classifier had been built) by linear least squares with nonnegativity constraints [35]. NMF parameter optimization has been done using parameter sweeps using the previously mentioned crossvalidation scheme to obtain performance measures; the criterion for parameter selection is chosen to be highest crossvalidated estimate of maximal accuracy out of 10 NMF restarts with random initializations. Having found the best set of parameters for each of the NMF algorithms, reported accuracies are

1) crossvalidated estimates of maximal accuracy (the criterion itself) out of 10 random restarts
2) mean and variance of accuracies of the entire crossvalidation scheme.

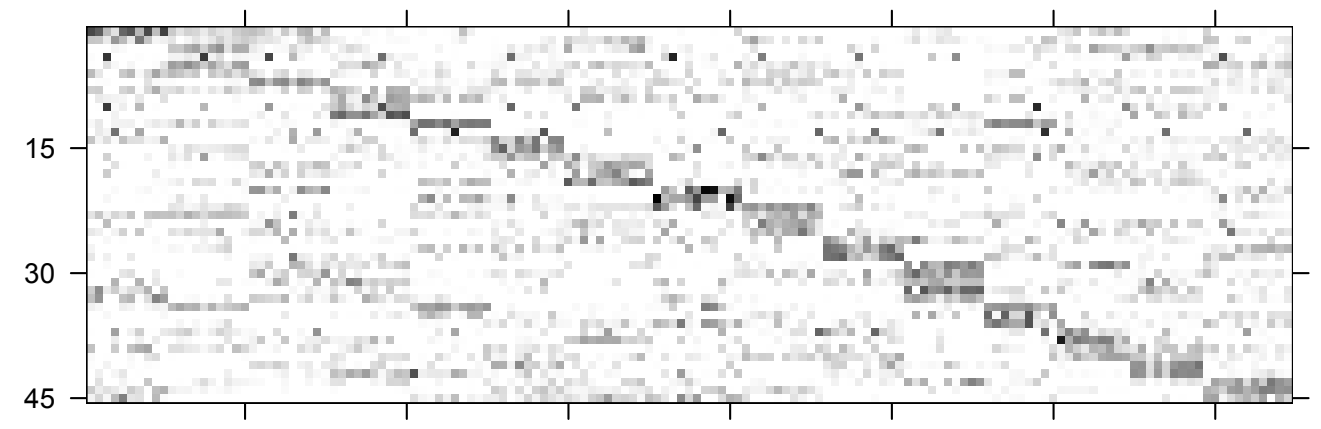

***nmf_gs***: $a^t_{vi} = 0.6$, $b^t_{vi} = 20$, $a^\lambda_{\text{small}} = 32$, $\boldsymbol{a}^\lambda_{\textbf{large}} = \mathbf{32}$, $b^\lambda = 1E6$

a)

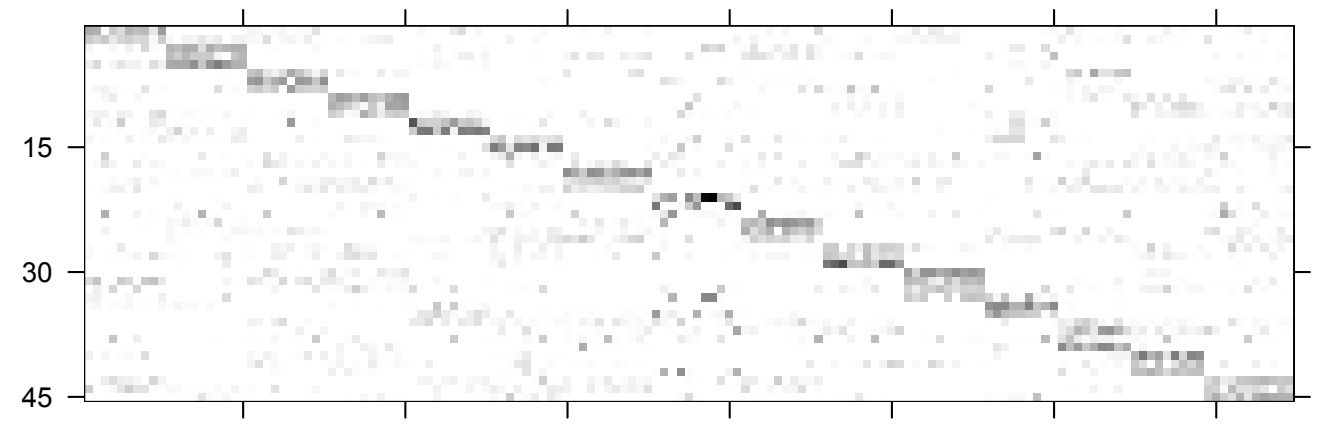

***nmf_gs***: $a^t_{vi} = 0.6$, $b^t_{vi} = 20$, $a^\lambda_{\text{small}} = 32$, $\boldsymbol{a}^\lambda_{\textbf{large}} = \mathbf{256}$, $b^\lambda = 1E6$

b)

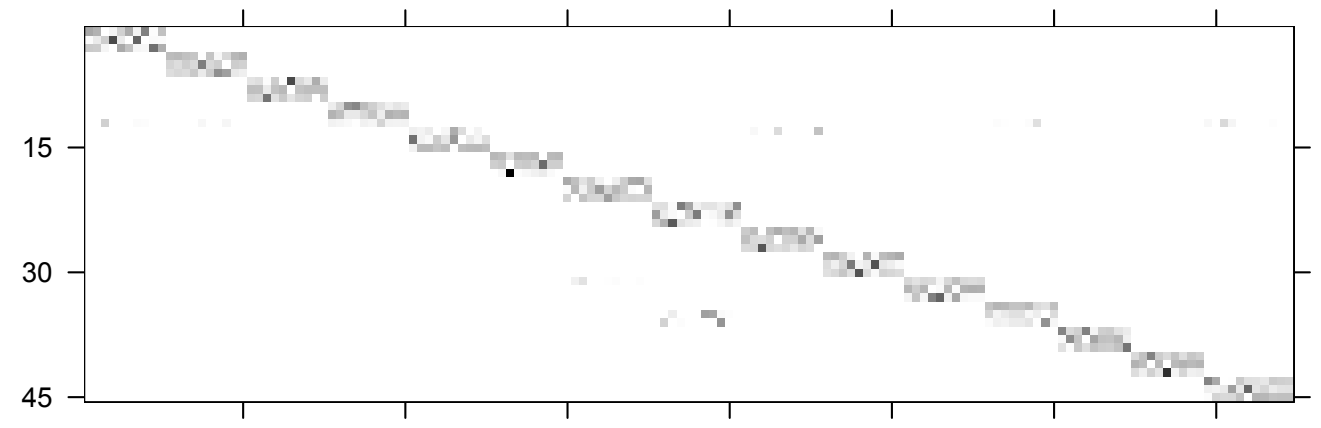

***nmf_gs***: $a^t_{vi} = 0.6$, $b^t_{vi} = 20$, $a^\lambda_{\text{small}} = 32$, $\boldsymbol{a}^\lambda_{\textbf{large}} = \mathbf{2048}$, $b^\lambda = 1E6$

c)

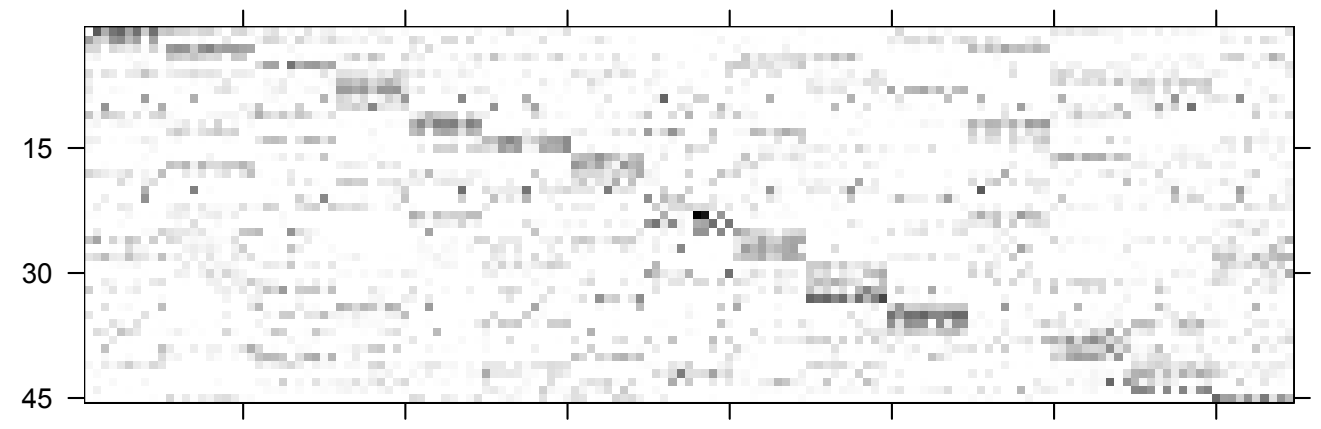

***nmf_vb***: $a^t_{vi} = 0.8$, $b^t_{vi} = 20$, $a^v_{i\tau} = 1$, $b^v_{i\tau} = 10$

d)

Fig. 2. Yale dataset, samples in feature space, i.e. the coefficient matrix $E_v$ , appropriatedly sorted: a), b) and c) *nmf_gs* with increasing parameter $a^\lambda_{\text{large}}$; d) *nmf_vb*, with parameters listed below their corresponding subimages. The darker the shade of grey, the higher the magnitude.

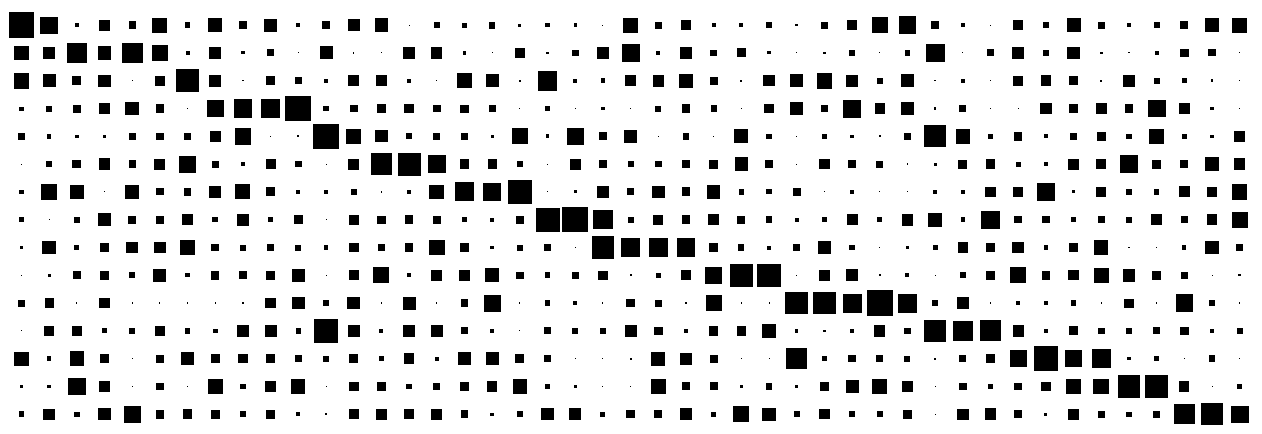

***nmf_gs***; $a^t_{vi} = 0.6$, $b^t_{vi} = 20$, $a^\lambda_{\text{small}} = 32$, $\boldsymbol{a}^\lambda_{\textbf{large}} = \mathbf{32}$, $b^\lambda = 1E6$

a)

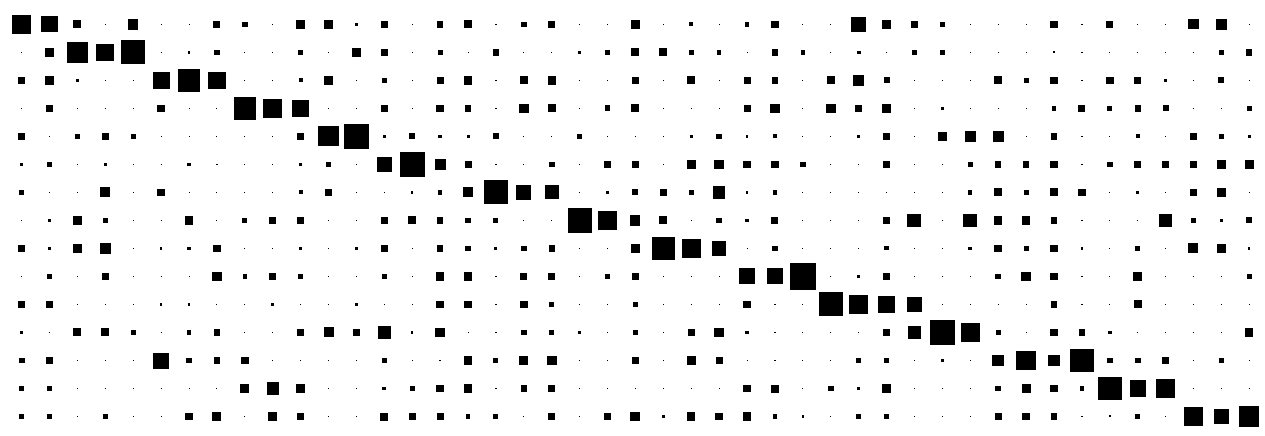

***nmf_gs***; $a^t_{vi} = 0.6$, $b^t_{vi} = 20$, $a^\lambda_{\text{small}} = 32$, $\boldsymbol{a}^\lambda_{\textbf{large}} = \mathbf{256}$, $b^\lambda = 1E6$

b)

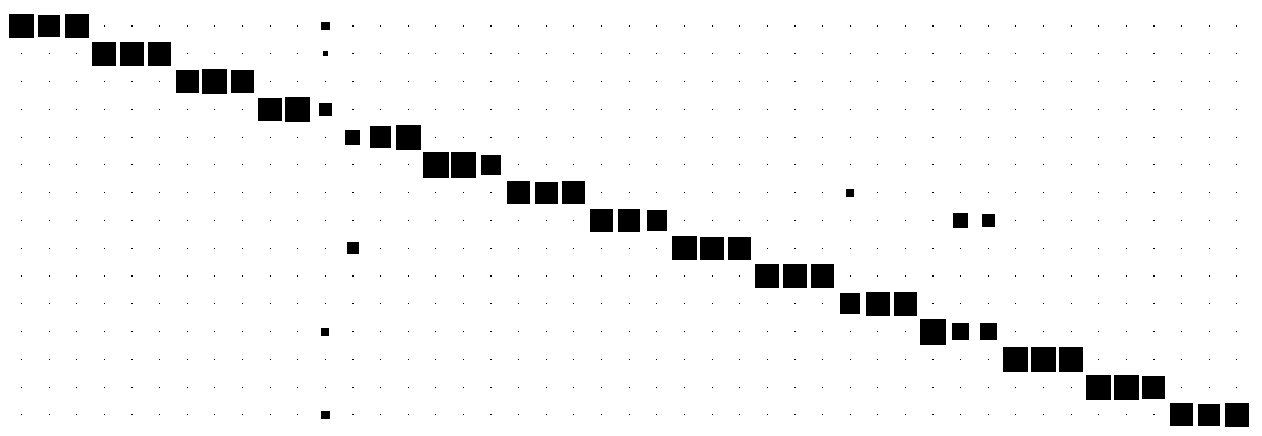

***nmf_gs***; $a^t_{vi} = 0.6$, $b^t_{vi} = 20$, $a^\lambda_{\text{small}} = 32$, $\boldsymbol{a}^\lambda_{\textbf{large}} = \mathbf{2048}$, $b^\lambda = 1E6$

c)

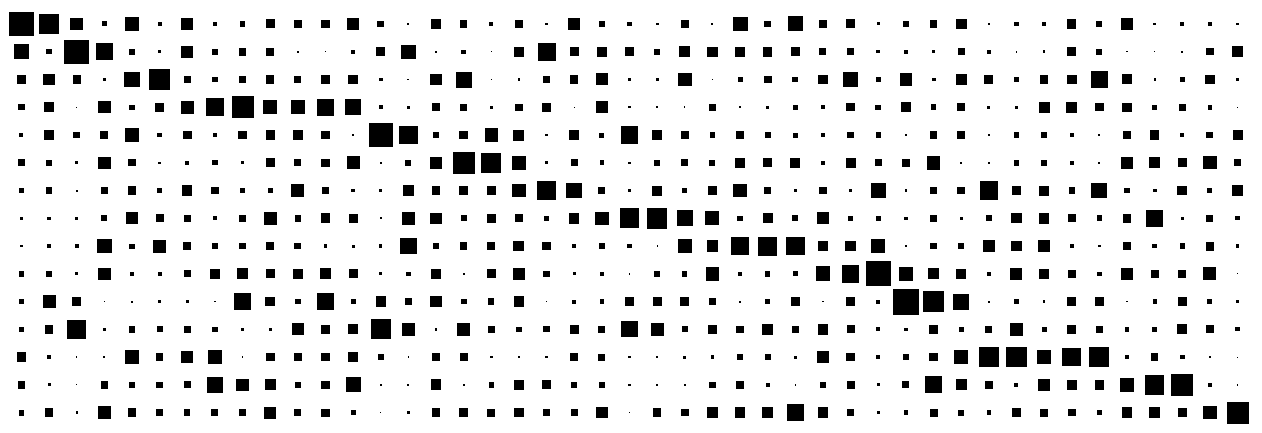

***nmf_vb***: $a^t_{vi} = 0.8$, $b^t_{vi} = 20$, $a^v_{i\tau} = 1$, $b^v_{i\tau} = 10$

d)

Fig. 3. Controlling prevalences of features in labels on Yale dataset - Hinton diagrams of normalized l1 norm of the coefficient matrix $E_v$ accumulated accross labels: a), b) and c) *nmf_gs* with increasing parameter $a^\lambda_{\text{large}}$; d) *nmf_vb*, with parameters listed below their corresponding subimages. Rows correspond to labels and columns to features.

Number of iterations for family of NMF algorithms has been set to 300. In case of PCA, data is projected to space of most significant principal components, and the crossvalidated estimate of the accuracy is reported. Decomposition algorithms used in the experiments for comparison will be referred to in short as

1) *pca*, principal component analysis [36]
2) *nmf_kl*, NMF with KL-divergence that includes a weighted penalty term to encourage sparsity in the right matrix [6]
3) *nmf_vb*, a variational Bayesian NMF which can model sparse decompositions [11]
4) *nmf_gs*, the proposed variational Bayesian NMF with group sparsity constraints.

The implementation *nmf_vb* is available from [37] and *nmf_kl* is short for *nmf_kl_sparse_v*, a part of NMFlib v0.1.3 library for MATLAB [38].

### 3.3 Results and Discussion

First, effect of hyperpriors $\alpha^\lambda_{ic}$ and $\beta^\lambda_{ic}$ on continuous indicators of presence of a feature accross group labels, $\lambda_{\text{ic}}$ are going to the examined. Specifically, it will be shown how features representative of specific group labels, i.e. which are prevalent in specific group labels, can be extracted from data this way.

Consider the hyperpriors $A^\lambda$ and $B^\lambda$ of forms

$$\boldsymbol{A}^\lambda = \left[\vec{a}_1^\lambda \;\; \dots \;\; \vec{a}_1^\lambda \;\; \vec{a}_2^\lambda \;\; \dots \;\; \vec{a}_2^\lambda \;\; \dots \;\; \vec{a}_C^\lambda \;\; \dots \;\; \vec{a}_C^\lambda\right]^{\mathrm{T}},$$

$$\vec{a}_1^\lambda = \left[a_{\mathrm{small}}^\lambda \;\; a_{\mathrm{large}}^\lambda \;\; \dots \;\; a_{\mathrm{large}}^\lambda \;\; a_{\mathrm{large}}^\lambda\right]^{\mathrm{T}},$$

$$\vec{a}_2^\lambda = \left[a_{\mathrm{large}}^\lambda \;\; a_{\mathrm{small}}^\lambda \;\; \dots \;\; a_{\mathrm{large}}^\lambda \;\; a_{\mathrm{large}}^\lambda\right]^{\mathrm{T}},$$

$$\vdots$$

$$\vec{a}_C^\lambda = \left[a_{\mathrm{large}}^\lambda \;\; a_{\mathrm{large}}^\lambda \;\; \dots \;\; a_{\mathrm{large}}^\lambda \;\; a_{\mathrm{small}}^\lambda\right]^{\mathrm{T}},$$

$$[\boldsymbol{B}^\lambda]_{ic} = b^\lambda, \qquad (11)$$

where $C$ denotes number of groups. If $a_{\mathrm{small}}^\lambda$ is smaller than $a_{\mathrm{large}}^\lambda$, the hyperprior is such that rows of $A^\lambda$ which contain $a_c^\lambda$ at some set of indices $S$ will bias the corresponding ${\lambda_{\mathrm{ic}}}^{-1}, i\epsilon S$ towards larger values, which hierarchically propagates to the related hidden factors $v_{\mathrm{i:}}, i\epsilon S$, giving their elements $v_{\mathrm{i\tau}}$ a sparse prior with large expected value if $z_\tau = c$ and with small expected value otherwise. Thus, such a hyperprior describes the tendency of the coefficients $v_{\mathrm{i\tau}}$ to be significantly large in samples belonging to a single group only. In the presented experiments number of such representative coefficients is chosen to be equally distributed among groups, i.e. each group will have the same number of representative features bound.

Behavior of prior (11) on the Yale dataset can be eyeballed from Fig. 2, Fig. 3 and Fig. 4, which all correspond to and visualize qualitatively typical mappings in the feature space, representative of several chosen parameter setups. Specifically, enforced by the prior (11) here are 3 representative features per each label; note that the dataset has 15 labels, which totals to 45 features. The number of samples in the training set is 149.

Fig. 2 presents samples in the feature space as heatmaps, having the samples sorted according to their labels and the coefficients according to their cumulative l1 norm per label averaged by number of samples per label. As a baseline, an example of *nmf_vb* decomposition is presented in Fig. 2d).

What is observed is that via magnitude of difference between $a_{\mathrm{large}}^\lambda$ and $a_{\mathrm{small}}^\lambda$ degree of mixing between specified group-prevalent features can be controlled. When $a_{\mathrm{large}}^\lambda = a_{\mathrm{small}}^\lambda$, group sparse decompositions are obtained with no prior which would bias distinct features to be prevalent accross specific groups, as depicted by Fig. 2a). The case when $a_{\mathrm{large}}^\lambda \gg a_{\mathrm{small}}^\lambda$ resembles the result of concatenating the NMF decompositions obtained for each label separately, i.e. each label has its group of features which are groupwise very strictly separated in terms of mixing, represented by Fig. 2c). Between those two extremes, a sweet spot for obtaining representation spaces with good discriminative properties may be found, a case which relates to Fig. 2b). Indicative of the justifiedness of this line of reasoning is the fact that Fig. 2b) has been obtained using parameters which yielded the best crossvalidated classification performance, as reported in Table 3.

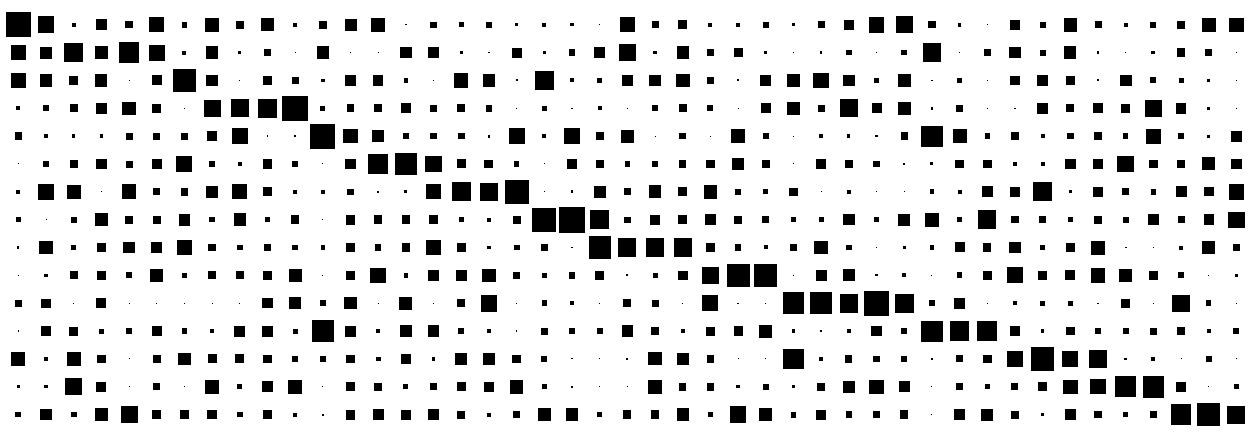

*nmf_gs*; $a_{vi}^t = 0.6$, $b_{vi}^t = 20$, $a_{\mathrm{small}}^\lambda = 32$, $\boldsymbol{a}_{\mathbf{large}}^\lambda = \mathbf{32}$, $b^\lambda = 1\mathrm{E}6$

a)

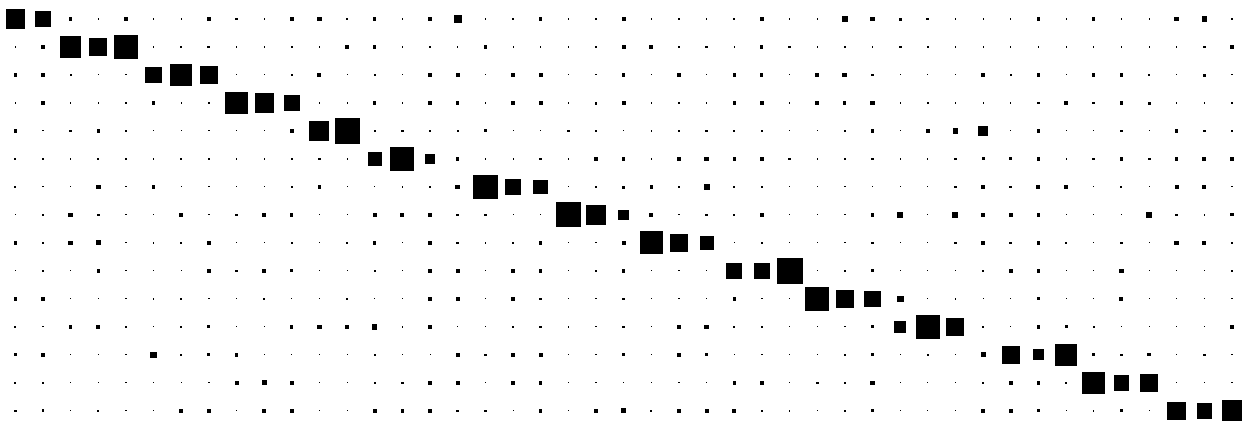

*nmf_gs*; $a_{vi}^t = 0.6$, $b_{vi}^t = 20$, $a_{\mathrm{small}}^\lambda = 32$, $\boldsymbol{a}_{\mathbf{large}}^\lambda = \mathbf{256}$, $b^\lambda = 1\mathrm{E}6$

b)

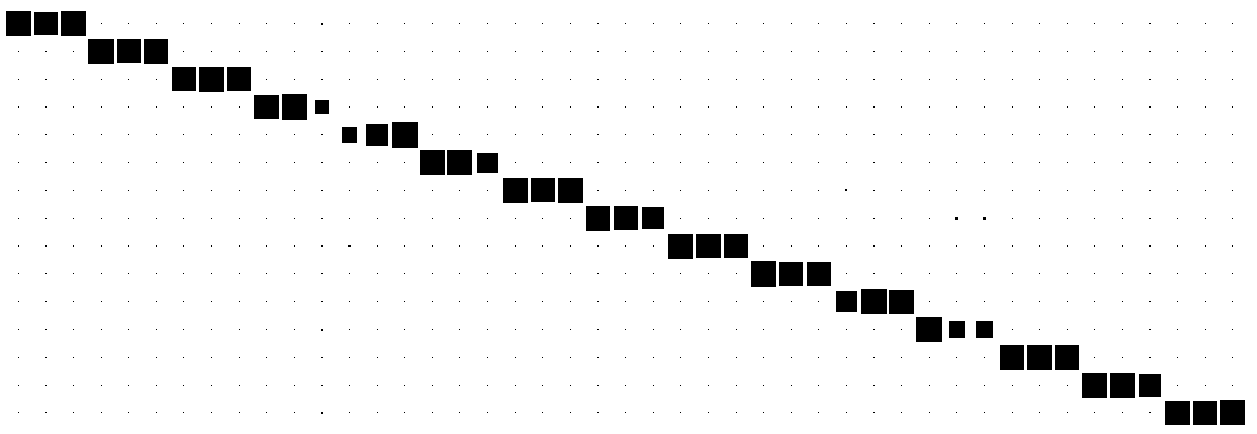

*nmf_gs*; $a_{vi}^t = 0.6$, $b_{vi}^t = 20$, $a_{\mathrm{small}}^\lambda = 32$, $\boldsymbol{a}_{\mathbf{large}}^\lambda = \mathbf{2048}$, $b^\lambda = 1\mathrm{E}6$

c)

Fig. 4. Yale dataset, Hinton diagrams of l1 norm of the reciprocal of group indicator variables ${E_\lambda}^{-1}$ accumulated accross labels for *nmf_gs* with increasing parameter $a_{\mathrm{large}}^\lambda$: a) $a_{\mathrm{large}}^\lambda = 32$, b) $a_{\mathrm{large}}^\lambda = 256$, c) $a_{\mathrm{large}}^\lambda = 2048$. Rows correspond to labels and columns to features.

From a different perspective, the same structural pattern over labels can be recognized using quantifiers as in Fig. 3 where, rather than presented directly accross samples, l1 norm of the coefficients has been accumulated accross samples having same label then normalized by cardinalities of the corresponding labels. Furthermore, in Fig. 4 diagrams of the same type but having the reciprocal of the group indicator matrix, ${E_\lambda}^{-1}$, as the target variable bear resemblance of a high degree to Fig. 3 a) b) and c), reason for which is that $E_\lambda$ directly specifies the prevalences of the features in each of the groups and propagates them hierarchically to variables $v_{\mathrm{i\tau}}$, according to (5).

Impact of choice of $a_{\mathrm{small}}^\lambda$ and $a_{\mathrm{large}}^\lambda$ on classification performance is illustrated on Fig. 6, Fig. 7 and Fig. 10 for Yale, ORL and JAFFE datasets, respectively. On Yale dataset, for a fixed $a_{\mathrm{small}}^\lambda$ as $a_{\mathrm{large}}^\lambda$ increases the accuracy improves, hitting a peak after which it begins to deteriorate, but not below the case when $a_{\mathrm{small}}^\lambda = a_{\mathrm{large}}^\lambda$. Qualitatively similar is the behavior on ORL dataset, but the accuracy improvement is more modest. On JAFFE dataset, improvement in the accuracy is notable but, more significantly, a pronounced droop when $\alpha^{\lambda}_{\text{large}} \gg \alpha^{\lambda}_{\text{small}}$ is present.

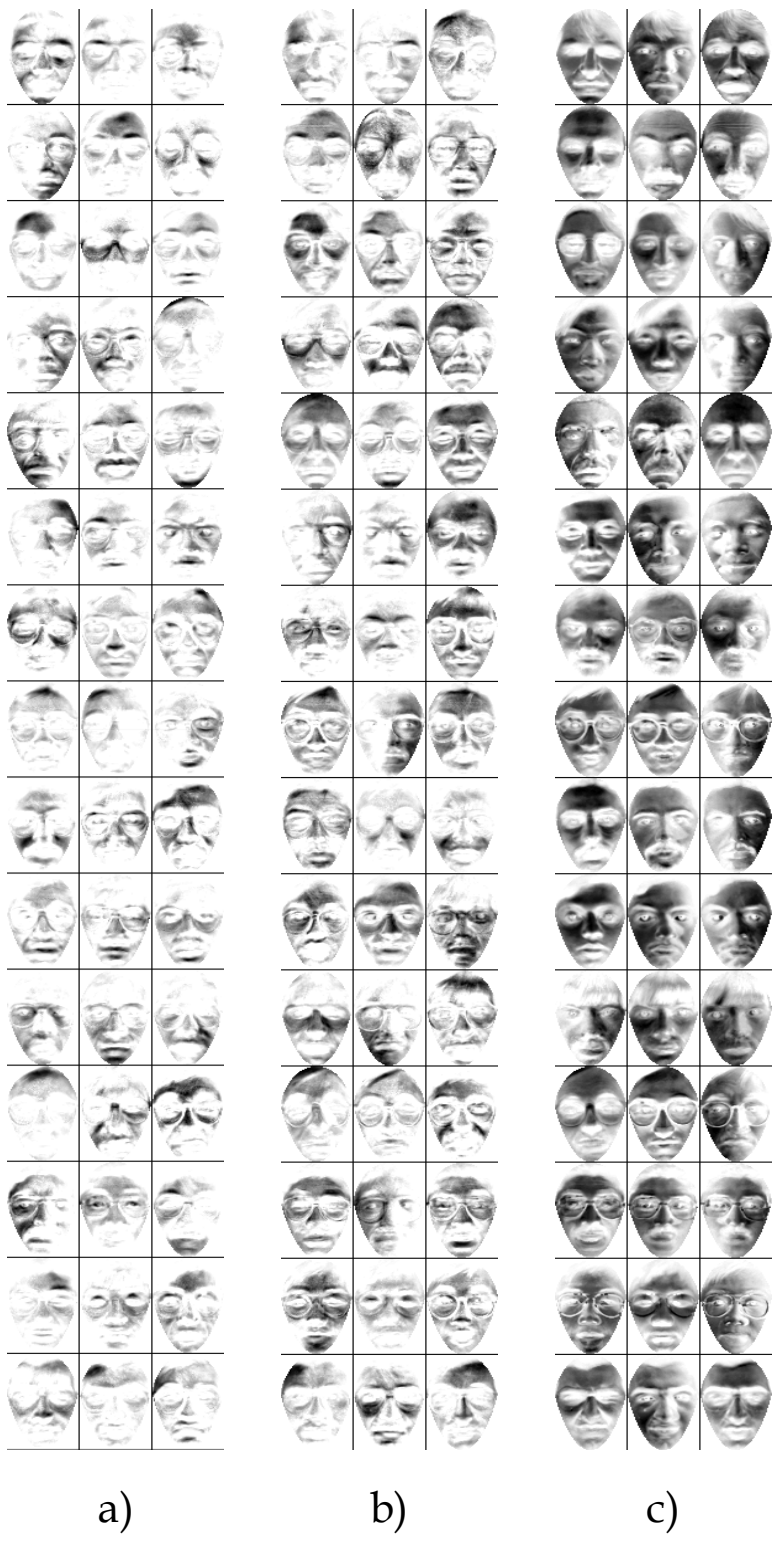

Fig. 5. Yale dataset, extracted features (in negative) for *nmf_gs* with increasing parameter $\alpha^{\lambda}_{\text{large}}$; a) $\alpha^{\lambda}_{\text{large}} = 32$, b) $\alpha^{\lambda}_{\text{large}} = 256$, c) $\alpha^{\lambda}_{\text{large}} = 2048$. Other parameters are $a^{t}_{vi} = 0.6$, $b^{t}_{vi} = 20$, $\alpha^{\lambda}_{\text{small}} = 32$, $b^{\lambda} = 1\text{E}6$.

Explanation for this effect is that, on JAFFE dataset, the droop is caused by a too restrictive mixing (Fig. 8c)), resulting in decompositions where same subjects with different expressions exhibit hardly any common features, i.e. expression-independent and subject-specific information is not shared between groups, and, consequently, the features are forced to be too holistic to extract expressions exclusively, (Fig. 9c)). The same effect is behind the behavior with the Yale dataset, however, Yale dataset is such that distances between same subjects having different expressions or configuration is smaller than distances between different subjects with same expressions or under same configuration, allowing satisfactory class discrimination even with such extremely holistic features (Fig. 5). To remind the reader, the goal on Yale dataset is subject recognition regardless of different expressions and configuration and on JAFFE face expression recognition regardless of the subject making it.
Experimental results on the Yale dataset are summarized in Table 3. Compared to the classification using *nmf_kl* and *nmf_vb*, improvement in the performance turned out to be significant when classification is performed in feature subspaces obtained by *nmf_gs*. It showed significantly higher average peak performance and higher average performance, with smaller variance also.

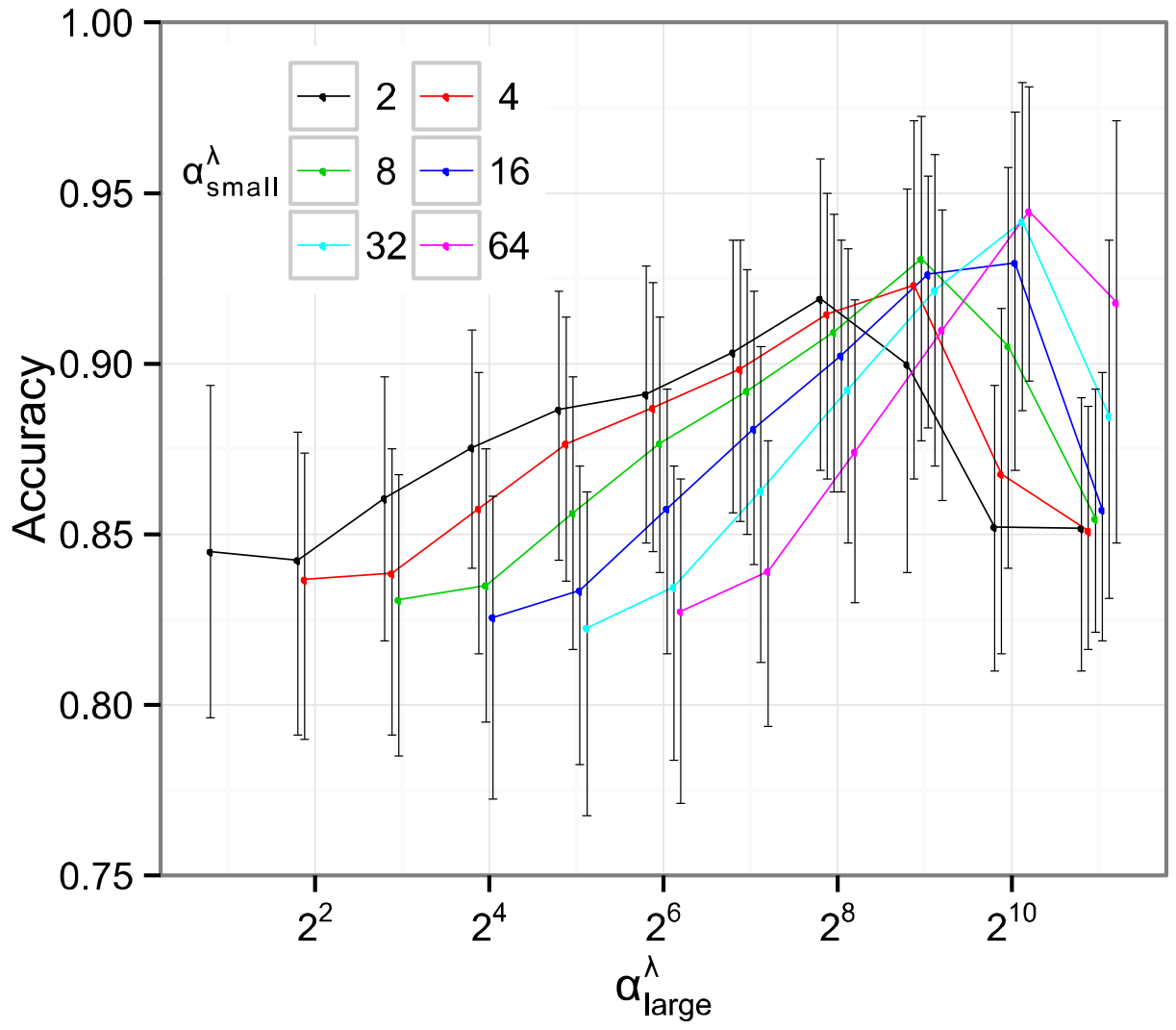

Fig. 6. Yale dataset, dependency of mean classification accuracy using *nmf_gs* on $\alpha^{\lambda}_{\text{small}}$ and $\alpha^{\lambda}_{\text{large}}$. Error bars represent crossvalidated estimates of maximal and minimal accuracies of a number of *nmf_gs* runs. Other parameters are fixed as $a^{t}_{vi} = 0.6$, $b^{t}_{vi} = 20$, $b^{\lambda} = 1\text{E}6$.

| Algorithm / Accuracy | *pca* | *nmf_kl* | *nmf_vb* | *nmf_gs* |
|---|---|---|---|---|
| maximum | 0.7987 | 0.9267 | 0.8875 | **0.9825** |
| mean ± variance | 0.7987 ± 0.0000 | 0.8690 ± 0.0054 | 0.8246 ± 0.0077 | **0.9417** ± 0.0036 |
| Subspace dimension | 73 | 120 | 60 | 45 |

Table 3. Classification results on the Yale dataset

On ORL dataset, as shown in Table 4, improvements are still observed, but to a far lesser extent. In the community, classification problem on the ORL dataset is known to lay on the easier side, as *pca* alone gives high accuracy. Regarding NMF algorithms, it can be concluded that sparsity constrains only are sufficient to give performance of high quality, leaving little space for improvement by group sparsity constraints.
Similar are the results on JAFFE dataset, presented in Table 5, but the improvement when using *gsNMF* is less marginal - compared to the results in Table 6 it can be seen that in this case *nmf_gs* in conjunction with 1-NN classifier can output classification results on the level of [39], where discrete wavelet transform with 2D linear discriminant analysis (LDA) is used to find features followed by classification using support vector machines with different kernel choices. Somewhat lower accuracies have been reported in [40] and [41]: experimental setup used in [40] consists of processing the samples by Gabor filtering, then sampling at fiducial points followed by PCA to get the features, finalized by LDA as classifier, and [41] uses Gabor filtering to get the features and two-layered perceptron for label discrimination. In [42] the authors use classifier based on Gaussian processes in the original pixel space.

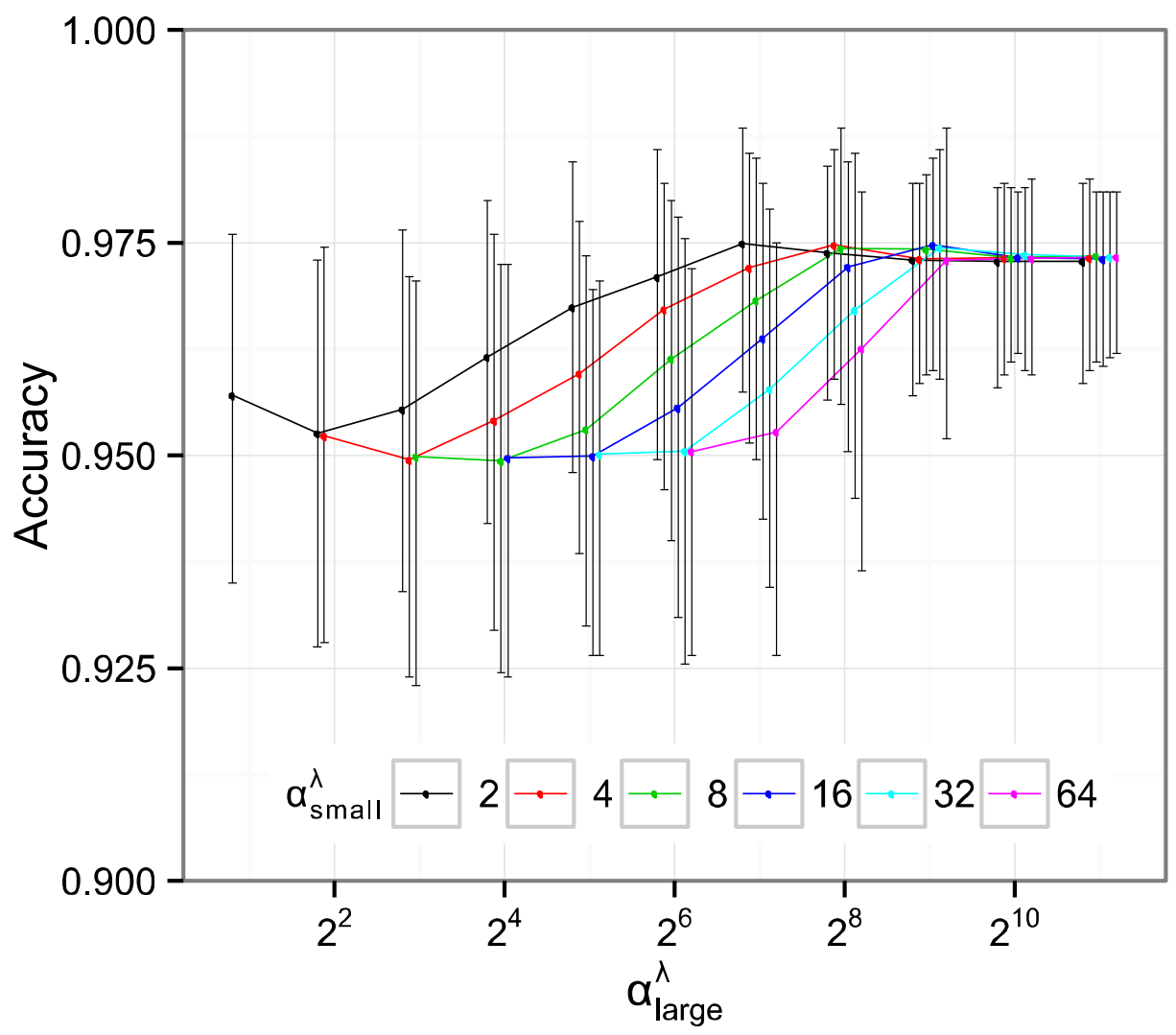

Fig. 7. ORL dataset, dependency of mean classification accuracy using *nmf_gs* on $a^{\lambda}_{\text{small}}$ and $a^{\lambda}_{\text{large}}$. Error bars represent crossvalidated estimates of maximal and minimal accuracies of a number of *nmf_gs* runs. Other parameters are fixed as $a^{t}_{vi} = 0.5$, $b^{t}_{vi} = 10, b^{\lambda} = 1E6$.

| Algorithm / Accuracy | *pca* | *nmf_kl* | *nmf_vb* | *nmf_gs* |
|---|---|---|---|---|
| maximum | 0.9605 | 0.9775 | 0.9780 | **0.9885** |
| mean ± variance | 0.9605 ± 0.0000 | 0.9529 ± 0.0013 | 0.9511 ± 0.0013 | **0.9750** ± 0.0006 |
| Subspace dimension | 67 | 140 | 60 | 160 |

Table 4. Classification results on the ORL dataset

Several other reported results on the same datasets are known to the author but are unfortunately of little use as the results have been evaluated nonuniformly accross publications.

From the practical point of view, however, even though peak performance of classification with *nmf_gs* is admirable, the problem of *a priori* selection of a *nmf_gs* decomposition which is bound to produce this peak remains open. This problem is not characteristic only of *nmf_gs*, but also of other NMF methods due to dependency of decompositions on initial values. Even though variational Bayes methodology allows calculation of variational bound which model comparisons can be made based upon, in the presented experiments the variational bound has been found to be uncorrelated with the classification accuracy, which is attributed to the fact that the classifier stands outside the Bayesian framework, i.e. that no objective directly connected with the classification had been embedded in the probabilistic model. Still, a solution always remains, which is to evaluate classification performance on a separate validation set and use it as an optimaliity indicator to determine which dictionary to select.

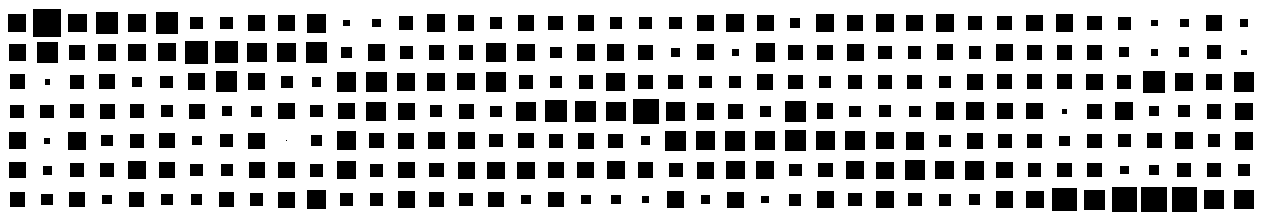

*nmf_gs*; $a^{t}_{vi} = 0.6$, $b^{t}_{vi} = 20$, $a^{\lambda}_{\text{small}} = 32$, $\boldsymbol{a^{\lambda}_{\text{large}}} = \mathbf{32}$, $b^{\lambda} = 1E6$

a)

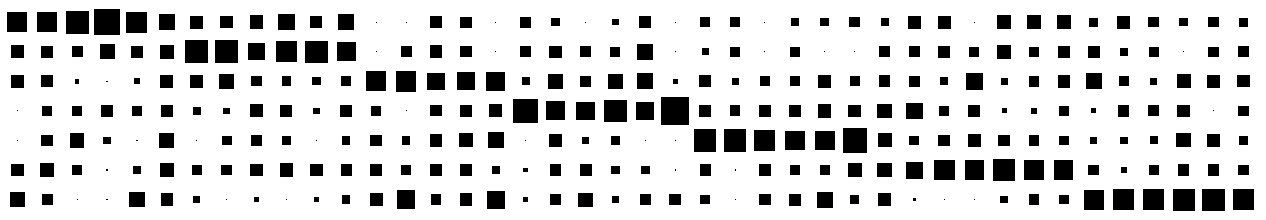

*nmf_gs*; $a^{t}_{vi} = 0.6$, $b^{t}_{vi} = 20$, $a^{\lambda}_{\text{small}} = 32$, $\boldsymbol{a^{\lambda}_{\text{large}}} = \mathbf{256}$, $b^{\lambda} = 1E6$

b)

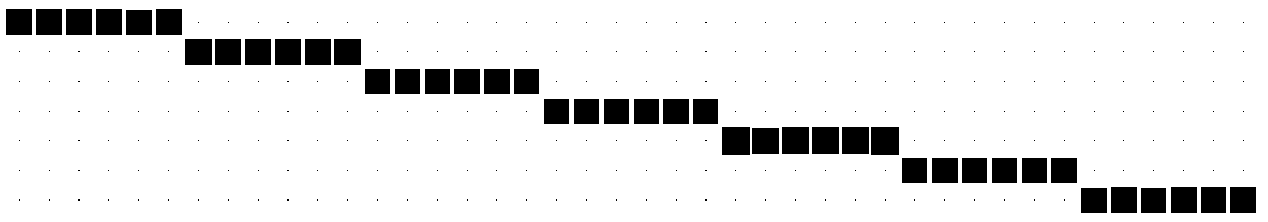

*nmf_gs*; $a^{t}_{vi} = 0.6$, $b^{t}_{vi} = 20$, $a^{\lambda}_{\text{small}} = 32$, $\boldsymbol{a^{\lambda}_{\text{large}}} = \mathbf{2048}$, $b^{\lambda} = 1E6$

c)

Fig. 8. Controlling prevalences of features in labels on JAFFE dataset - Hinton diagrams of normalized l1 norm of the coefficient matrix $E_v$ accumulated accross labels for *nmf_gs* with increasing parameter $a^{\lambda}_{\text{large}}$: a) $a^{\lambda}_{\text{large}} = 32$, b) $a^{\lambda}_{\text{large}} = 256$, c) $a^{\lambda}_{\text{large}} = 2048$. Rows correspond to labels and columns to features.

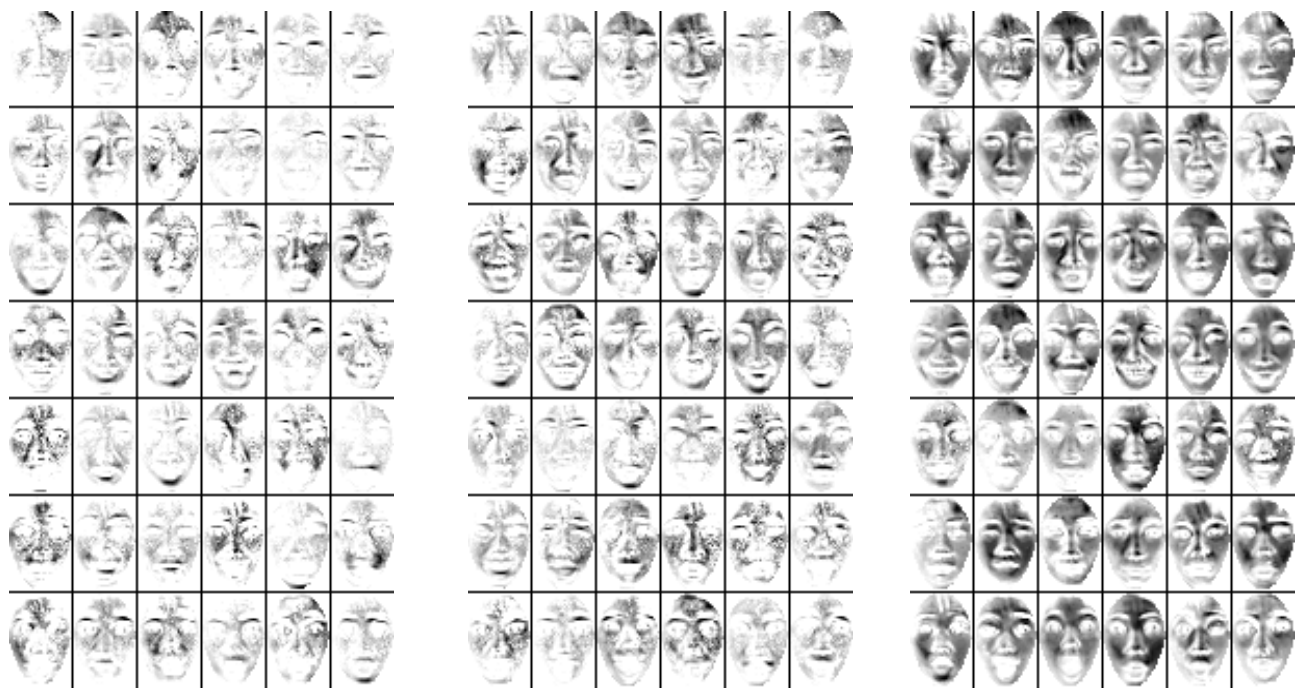

a) b) c)

Fig. 9. JAFFE dataset, extracted features (in negative) for *nmf_gs* with increasing parameter $a^{\lambda}_{\text{large}}$: a) $a^{\lambda}_{\text{large}} = 32$, b) $a^{\lambda}_{\text{large}} = 256$, c) $a^{\lambda}_{\text{large}} = 2048$. Other parameters are $a^{t}_{vi} = 0.6$, $b^{t}_{vi} = 20$, $a^{\lambda}_{\text{small}} = 32$, $b^{\lambda} = 1E6$.

## 4 Conclusion

A probabilistic formulation of NMF with group sparsity constraints has been layed out with an efficient variational Bayesian algorithm for approximate learning of the model parameters. It has been shown how prevalence of specific features accross groups and the degree of their mixing between groups can be controlled. Having identity-mapped the class labels to a more general notion of groups, the presented model has been utilized as a supervised feature extractor in face recognition and facial expression recognition applications and beneficial effects of such decomposition subspaces on classification performance have been observed.

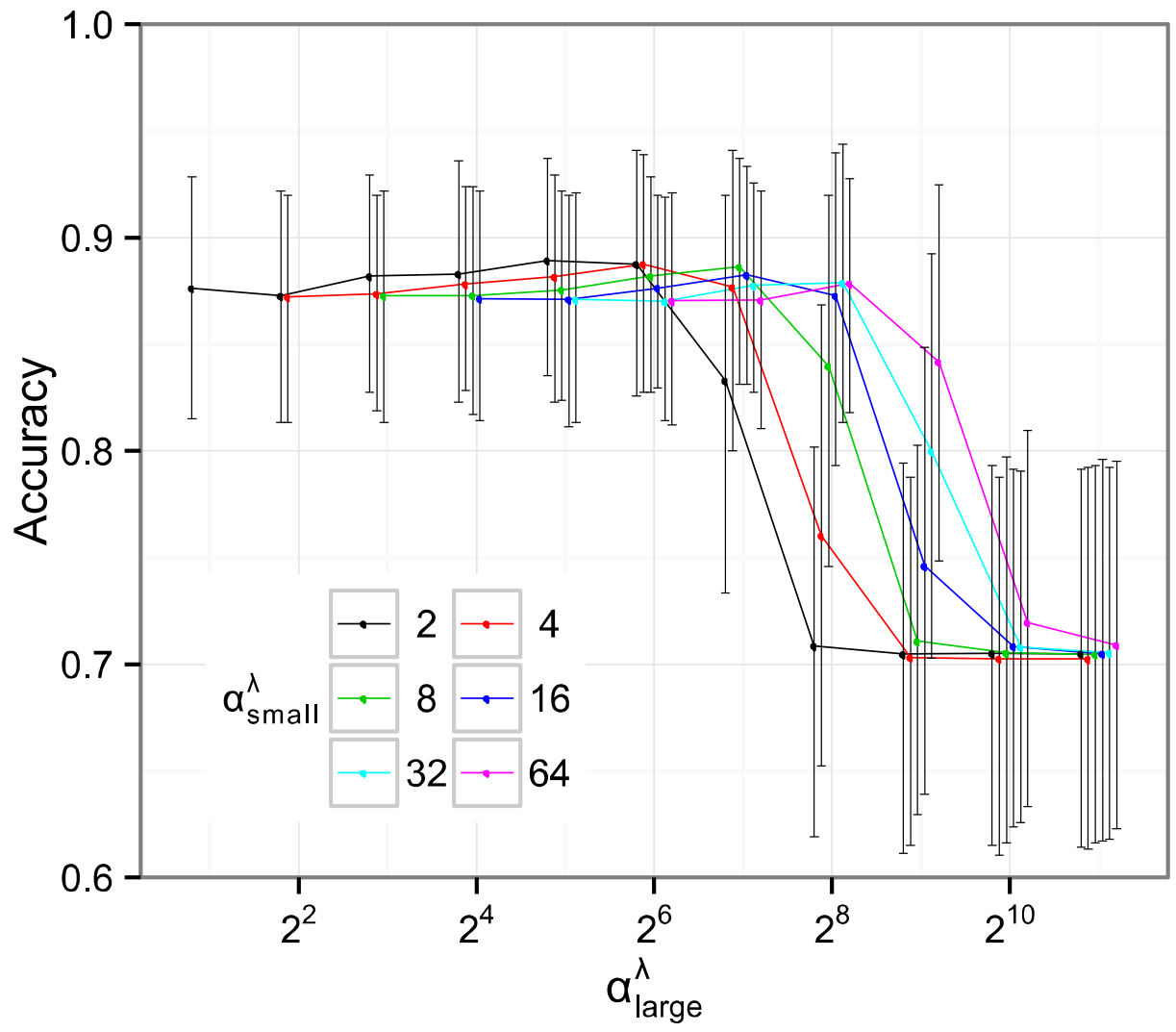

Fig. 10. JAFFE dataset, dependency of mean classification accuracy using *nmf_gs* on $a^{\lambda}_{\mathrm{small}}$ and $a^{\lambda}_{\mathrm{large}}$. Error bars represent crossvalidated estimates of maximal and minimal accuracies of a number of *nmf_gs* runs. Other parameters are fixed as $a^{t}_{vi} = 0.5$, $b^{t}_{vi} = 20$, $b^{\lambda} = 1E6$.

To contribute to community where easily reproducible research is appreciated, the implementation used in the experiments is made publicly available (which unfortunately is not the case still too often).

| Algorithm / Accuracy | *Pca* | *nmf_kl* | *nmf_vb* | *nmf_gs* |
|---|---|---|---|---|
| maximum | 0.9067 | 0.9267 | 0.9295 | **0.9438** |
| mean ± variance | **0.9067** ± 0.0000 | 0.8690 ± 0.0054 | 0.8597 ± 0.0063 | 0.8789 ± 0.0051 |
| Subspace dimension | 18 | 28 | 14 | 42 |

Table 5. Classification results on the JAFFE dataset

| Method | Shih et al. [39] | Lyons et al. [40] | Zhang et al. [41] | Cheng et al. [42] |
|---|---|---|---|---|
| Accuracy | **0.9413** | 0.92 | 0.901 | 0.8689 |

Table 6. Relevant classification results on the JAFFE dataset reported in the literature where accuracy has been obtained by 10-fold crossvalidation

Future work on the presented subject includes pursuing modifications of the group sparse formulation which would work in semi-supervised settings. The model should ideally allow efficient inference and learning of the labels of unlabeled data, avoiding sampling techniques for execution speed if possible.

## Acknowledgments

This work was supported by the Croatian Ministry of Science, Education and Sports through the project "Computational Intelligence Methods in Measurement Systems", No. 098-0982560-2565

**Ivan Ivek** received his B.S. in electrical engineering from the Faculty of Electrical Engineering and Computing, University of Zagreb, Zagreb, Croatia in 2007.
He joined the Ruđer Boškovic Institute in 2008, where his current position is Research Assistant for the Computational Intelligence Methods in Measurement Systems project. Currently, he is pursuing his Ph.D. in electronics at the Faculty of Electrical Engineering and Computing, University of Zagreb, Croatia.

## Appendix A - Probability Density Functions

By 1 (column) vector of ones and by $\Psi(.)$ elementwise digamma function, $\Psi(x) = \frac{\mathrm{d}}{\mathrm{dx}} \log \Gamma(x)$.

| |
|---|
| $Poisson(x\|\lambda) = \exp(-\lambda + x\log(\lambda) - \log\Gamma(x+1))$,<br>with $\lambda > 0$<br>$\langle x \rangle = \lambda;$ |
| $Gamma(x\|a,b) = \exp\left(-\frac{1}{b}x + (a-1)\log x - a\log b - \log\Gamma(a)\right)$,<br>with $a > 0, b > 0;$<br>$\langle x \rangle = ab,$<br>$\langle \log x \rangle = \Psi(a) + \log b;$<br>$\mathcal{H}(x) = -(a-1)\psi(a) + \log b + a + \log\Gamma(a);$ |
| $Exponential(x\|b) = Gamma(x\|1,b);$ |
| $Multinomial\left(\begin{bmatrix} x_1 \\ \vdots \\ x_C \end{bmatrix} \middle\| s, \begin{bmatrix} p_1 \\ \vdots \\ p_C \end{bmatrix}\right) = \delta(s - \sum_i x_i)\exp(\log\Gamma(s+1) + \sum_i(x_i \log p_i - \log\Gamma(x_i+1))),$<br>with $\sum_i p_i = 1, \sum_i x_i = s;$<br>$\begin{bmatrix} \langle \log x_1 \rangle \\ \vdots \\ \langle \log x_C \rangle \end{bmatrix} = s \begin{bmatrix} p_1 \\ \vdots \\ p_C \end{bmatrix};$<br>$\mathcal{H}(\vec{x}) = -\log\Gamma(s+1) - \sum_i \langle x_i \rangle \log p_i + \sum_i \langle \log\Gamma(x_i+1) \rangle - \langle \log\delta(s - \sum_i x_i) \rangle;$ |
| $Dirichlet\left(\begin{bmatrix} x_1 \\ \vdots \\ x_C \end{bmatrix} \| \begin{bmatrix} u_1 \\ \vdots \\ u_C \end{bmatrix}\right) = \exp\left(\begin{bmatrix} u_1 - 1 \\ \vdots \\ u_C - 1 \end{bmatrix}^T \begin{bmatrix} \log x_1 \\ \vdots \\ \log x_C \end{bmatrix} + \log\Gamma(\sum_c^C u_c) - \sum_c^C \Gamma(u_i)\right),$<br>with $\sum_c x_c = 1, u_c > 0;$<br>$\begin{bmatrix} \langle \log x_1 \rangle \\ \vdots \\ \langle \log x_C \rangle \end{bmatrix} = \begin{bmatrix} \Psi(u_1) \\ \vdots \\ \Psi(u_C) \end{bmatrix} - \Psi(\sum_c^C u_c);$<br>$\mathcal{H}\left(x \begin{bmatrix} x_1 \\ \vdots \\ x_C \end{bmatrix}\right) = -\log\Gamma(\sum_{c=1}^C u_c) + \Gamma(\sum_{c=1}^C u_c) + (\sum_{c=1}^C u_c - C)\psi(\sum_{c=1}^C u_c) - \sum_{c=1}^C (u_c - 1)\psi(u_c);$ |
| $Discrete(x\| \begin{bmatrix} \log p_1 \\ \vdots \\ \log p_C \end{bmatrix}) = \exp\left(\begin{bmatrix} \log p_1 \\ \vdots \\ \log p_C \end{bmatrix}^T \begin{bmatrix} \delta(x-1) \\ \vdots \\ \delta(x-C) \end{bmatrix}\right),$<br>with $\sum_i p_i = 1;$<br>$\begin{bmatrix} \langle \delta(x-1) \rangle \\ \vdots \\ \langle \delta(x-C) \rangle \end{bmatrix} = \begin{bmatrix} p_1 \\ \vdots \\ p_C \end{bmatrix}^T;$<br>$\mathcal{H}(x) = -\sum_{c=1}^C p_c \log p_c;$ |

## Appendix B - Variational Bound

Applying (2) on the presented model, the bound is expanded as

$$
\begin{aligned}
\mathcal{L}(q)^{(n)} = & \sum_{\nu,\tau} \langle \log p(x_{\nu\tau}|s_{\nu:\tau})\rangle_{q(s_{\nu:\tau})^{(n)}} \\
& + \sum_{\nu,i,\tau} \langle \log p(s_{\nu i\tau}|t_{\nu i}, v_{i\tau})\rangle_{q(s_{\nu i\tau})^{(n)} q(t_{\nu i})^{(n)} q(v_{i\tau})^{(n)}} + \sum_{\nu,i,\tau} \mathcal{H}(q(s_{\nu i\tau})^{(n)}) \\
& + \sum_{\nu,i} \langle \log p(t_{\nu i}|a_{\nu i}^{t}, b_{\nu i}^{t})\rangle_{q(t_{\nu i})^{(n)}} + \sum_{\nu,i} \mathcal{H}(q(t_{\nu i})^{(n)}) \\
& + \sum_{i,\tau} \langle \log p(v_{i\tau}|z_{\tau}, \lambda_{i:})\rangle_{q(v_{i\tau})^{(n)} q(z_{\tau})^{(n)} q(\lambda_{i:})^{(n)}} + \sum_{i,\tau} \mathcal{H}(q(v_{i\tau})^{(n)}) \\
& + \sum_{i,c} \langle \log p(\lambda_{ic}|a_{ic}^{\lambda}, b_{ic}^{\lambda})\rangle_{q(\lambda_{ic})^{(n)}} + \sum_{i,c} \mathcal{H}(q(\lambda_{ic})^{(n)}) \\
& + \sum_{\tau,c} \langle \log p(z_{\tau}|\pi_{\tau:})\rangle_{q(z_{\tau})^{(n)} q(\pi_{\tau:})^{(n)}} + \sum_{\tau} \mathcal{H}(q(z_{\tau})^{(n)}) \\
& + \sum_{\tau} \langle p\,([\pi_{\tau 1}, \dots, \pi_{\tau C}]^{T} | [u_{\tau 1}, \dots, u_{\tau C}]^{T})\rangle_{q(\pi_{\tau:})^{(n)}} + \sum_{\tau} \mathcal{H}(q([\pi_{\tau 1}, \dots, \pi_{\tau C}]^{T})^{(n)}).
\end{aligned}
$$

With update expressions ordered as in Appendix C and bound calculated where marked, substitutions

$$
\begin{aligned}
& \Psi\left(\alpha_{\nu i}^{t\,(n)}\right) = \langle \log t_{\nu i}\rangle^{(n+1)} - \log \beta_{\nu i}^{t\,(n)} \\
& \Psi\left(\alpha_{i\tau}^{v\,(n)}\right) = \langle \log v_{i\tau}\rangle^{(n+1)} - \log \beta_{i\tau}^{v\,(n)} \\
& \alpha_{\nu i}^{t\,(n)} = a_{\nu i}^{t} + \sum_{\tau} \langle s_{\nu i\tau}\rangle^{(n)} \\
& \alpha_{i\tau}^{v\,(n)} = 1 + \sum_{\nu} \langle s_{\nu i\tau}\rangle^{(n)}
\end{aligned}
$$

can be used to eliminate more costly evaluations of digamma function. Then, the bound adopts the form

$$
\begin{aligned}
\mathcal{L}(q)^{(n)} = & -\sum_{\nu,\tau} \left( \boldsymbol{E}_{\boldsymbol{t}}^{(n)} \boldsymbol{E}_{\boldsymbol{v}}^{(n)} + \log \Gamma(\boldsymbol{X}+1) + \log\left(\exp \boldsymbol{L}_{\boldsymbol{t}}^{(n)} * \exp \boldsymbol{L}_{\boldsymbol{v}}^{(n)}\right) \right) \\
& + \sum_{\nu,\tau} \left( -\boldsymbol{X}.* \left( \left(\exp \boldsymbol{L}_{\boldsymbol{t}}^{(n)} .* \boldsymbol{L}_{\boldsymbol{t}}^{(n)}\right) * \exp \boldsymbol{L}_{\boldsymbol{v}}^{(n)} + \exp \boldsymbol{L}_{\boldsymbol{t}}^{(n)} * \left(\exp \boldsymbol{L}_{\boldsymbol{v}}^{(n)} .* \boldsymbol{L}_{\boldsymbol{v}}^{(n)}\right) ./ \left(\exp \boldsymbol{L}_{\boldsymbol{t}}^{(n)} * \exp \boldsymbol{L}_{\boldsymbol{v}}^{(n)}\right) \right) \right) \\
& + \sum_{\nu,i} \left( -1./\boldsymbol{B}^{\boldsymbol{t}}.* \boldsymbol{E}_{\boldsymbol{t}}^{(n)} - \log \Gamma(\boldsymbol{A}^{\boldsymbol{t}}) - \boldsymbol{A}^{\boldsymbol{t}}.* \log \boldsymbol{B}^{\boldsymbol{t}} + \boldsymbol{A}_{\boldsymbol{t}}{}^{(n)}.* \left(\log \boldsymbol{B}_{\boldsymbol{t}}{}^{(n)} + 1\right) + \log \Gamma\left(\boldsymbol{A}_{\boldsymbol{t}}{}^{(n)}\right) \right) \\
& + \sum_{i,\tau} \left( -\boldsymbol{E}_{\boldsymbol{\lambda}}^{(n)} * \boldsymbol{\Delta}^{(n)^{T}} .* \boldsymbol{E}_{\boldsymbol{v}}^{(n)} + \log\left(\boldsymbol{E}_{\boldsymbol{\lambda}}^{(n)} * \boldsymbol{\Delta}^{(n)^{T}}\right) - \boldsymbol{A}^{\boldsymbol{v}}.* \log \boldsymbol{B}^{\boldsymbol{v}} + \boldsymbol{A}_{\boldsymbol{v}}{}^{(n)}.* \left(\log\ \boldsymbol{B}_{\boldsymbol{v}}{}^{(n)} + 1\right) + \log \Gamma\left(\boldsymbol{A}_{\boldsymbol{v}}{}^{(n)}\right) \right) \\
& + \sum_{i,c} \left( -1./\boldsymbol{B}^{\boldsymbol{\lambda}}.* \boldsymbol{E}_{\boldsymbol{\lambda}}^{(n)} + \left(\boldsymbol{A}^{\boldsymbol{\lambda}} - 1\right).* \boldsymbol{L}_{\boldsymbol{\lambda}}^{(n)} - \boldsymbol{A}^{\boldsymbol{\lambda}}.* \log \boldsymbol{B}^{\boldsymbol{\lambda}} - \log \Gamma\left(\boldsymbol{A}^{\boldsymbol{\lambda}}\right) \right) \\
& + \sum_{i,c} \left( \left(1 - \boldsymbol{A}_{\boldsymbol{\lambda}}{}^{(n)}\right) * \psi\left(\boldsymbol{A}_{\boldsymbol{\lambda}}{}^{(n)}\right) + \log \boldsymbol{B}_{\boldsymbol{\lambda}}{}^{(n)} + \boldsymbol{A}_{\boldsymbol{\lambda}}{}^{(n)} + \log \Gamma\left(\boldsymbol{A}_{\boldsymbol{\lambda}}{}^{(n)}\right) \right) \\
& + \sum_{\tau,c} (\boldsymbol{U} - 1).* \boldsymbol{\Pi}^{(n)^{T}} + \sum_{\tau} \left(\boldsymbol{U} * \vec{1} - \textstyle\sum_{c} 1\right).* \psi\left(\boldsymbol{U} * \vec{1}\right) - \sum_{\tau,c} (\boldsymbol{U} - 1).* \psi(\boldsymbol{U}) \quad , \qquad (14)
\end{aligned}
$$

where by $\sum_{c} 1$ the number of columns of $U$ is denoted.

## Appendix C - Summary of the Learning Algorithm

Inputs: $\boldsymbol{X}, \boldsymbol{A}^{\boldsymbol{t}}, \boldsymbol{B}^{\boldsymbol{t}}, \boldsymbol{A}^{\boldsymbol{\lambda}}, \boldsymbol{B}^{\boldsymbol{\lambda}}, \boldsymbol{U}$

Initialize (randomly):

$$\boldsymbol{E}_{\boldsymbol{t}}^{(0)}, \boldsymbol{L}_{\boldsymbol{t}}^{(0)}, \boldsymbol{E}_{\boldsymbol{v}}^{(0)} \boldsymbol{L}_{\boldsymbol{v}}^{(0)}, \boldsymbol{\Sigma}_{\boldsymbol{t}}^{(0)}, \boldsymbol{\Sigma}_{\boldsymbol{v}}^{(0)}, \boldsymbol{\Delta}^{(0)}, \boldsymbol{E}_{\boldsymbol{\lambda}}^{(0)}, \boldsymbol{L}_{\boldsymbol{\lambda}}^{(0)}, \boldsymbol{\Pi}^{(0)}, \boldsymbol{\Upsilon}^{(0)}$$

$$j = [j_1, \dots, j_i, \dots]^T, j_i = i$$

$n = 0;$

Repeat:

$$\boldsymbol{\xi}^{(n+1)} = \boldsymbol{X}./((\exp \boldsymbol{L}_{\boldsymbol{t}}^{(n)}) * (\exp \boldsymbol{L}_{\boldsymbol{v}}^{(n)}))$$
$$\boldsymbol{\Sigma}_{\boldsymbol{v}}^{(n+1)} = \exp \boldsymbol{L}_{\boldsymbol{v}}^{(n)} .* (\exp \boldsymbol{L}_{\boldsymbol{t}}^{(n)})^T * \boldsymbol{\xi}^{(n)})$$
$$\boldsymbol{\Sigma}_{\boldsymbol{t}}^{(n+1)} = \exp \boldsymbol{L}_{\boldsymbol{t}}^{(n)} .* (\boldsymbol{\xi}^{(n)} * (\exp \boldsymbol{L}_{\boldsymbol{v}}^{(n)})^T)$$

$$\boldsymbol{A}_{\boldsymbol{t}}^{(n)} = 1.* \boldsymbol{A}^{\boldsymbol{t}} + \boldsymbol{\Sigma}_{\boldsymbol{t}}^{(n)}$$
$$\boldsymbol{B}_{\boldsymbol{t}}^{(n)} = 1./(1./\boldsymbol{B}^{\boldsymbol{t}} + \boldsymbol{1} * \boldsymbol{E}_{\boldsymbol{v}}^{(n)^T})$$
$$\boldsymbol{E}_{\boldsymbol{t}}^{(n+1)} = \boldsymbol{A}_{\boldsymbol{t}}^{(n)} .* \boldsymbol{B}_{\boldsymbol{t}}^{(n)}$$

$$\boldsymbol{A}_{\boldsymbol{v}}^{(n)} = 1 + \boldsymbol{\Sigma}_{\boldsymbol{v}}^{(n)}$$
$$\boldsymbol{B}_{\boldsymbol{v}}^{(n)} = 1./(\boldsymbol{E}_{\boldsymbol{\lambda}}^{(n)} * \boldsymbol{\Delta}^{(n)^{\mathrm{T}}} + \boldsymbol{E}_{\boldsymbol{t}}^{(n)^T} * \boldsymbol{1})$$
$$\boldsymbol{E}_{\boldsymbol{v}}^{(n+1)} = \boldsymbol{A}_{\boldsymbol{v}}^{(n)} .* \boldsymbol{B}_{\boldsymbol{v}}^{(n)}$$

Optional: calculate bound according to (14)

$$\boldsymbol{L}_{\boldsymbol{t}}^{(n+1)} = \Psi(\boldsymbol{A}_{\boldsymbol{t}}^{(n)}) + \log \boldsymbol{B}_{\boldsymbol{t}}^{(n)}$$
$$\boldsymbol{L}_{\boldsymbol{v}}^{(n+1)} = \Psi(\boldsymbol{A}_{\boldsymbol{v}}^{(n)}) + \log \boldsymbol{B}_{\boldsymbol{v}}^{(n)}$$

$$\boldsymbol{A}_{\boldsymbol{\lambda}}^{(n)} = \boldsymbol{A}^{\boldsymbol{\lambda}} + \boldsymbol{\Delta}^{(n)} * \boldsymbol{1}$$
$$\boldsymbol{B}_{\boldsymbol{\lambda}}^{(n)} = 1./\left(1./\boldsymbol{B}^{\boldsymbol{\lambda}} + \boldsymbol{E}_{\boldsymbol{v}}^{(n)} * \boldsymbol{\Delta}^{(n)}\right)$$
$$\boldsymbol{E}_{\boldsymbol{\lambda}}^{(n+1)} = \boldsymbol{A}_{\boldsymbol{\lambda}}^{(n)} .* \boldsymbol{B}_{\boldsymbol{\lambda}}^{(n)}$$
$$\boldsymbol{L}_{\boldsymbol{\lambda}}^{(n+1)} = \Psi(\boldsymbol{A}_{\boldsymbol{\lambda}}^{(n)}) + \log \boldsymbol{B}_{\boldsymbol{\lambda}}^{(n)}$$

$$\boldsymbol{P}^{(n)} = \boldsymbol{\Pi}^{(n)} + \boldsymbol{E}_{\boldsymbol{t}}^{(n)^{\mathrm{T}}} * \boldsymbol{E}_{\boldsymbol{\lambda}}^{(n)} + \boldsymbol{1} * \boldsymbol{L}_{\boldsymbol{\lambda}}^{(n)}$$
$$\boldsymbol{\Delta}^{(n+1)} = \boldsymbol{P}^{(n)}./(\boldsymbol{P}^{(n)} * \boldsymbol{1})$$

$$\boldsymbol{\Upsilon}^{(n)} = \boldsymbol{U}^{(n)} + \boldsymbol{\Delta}^{(n)}$$
$$\boldsymbol{\Pi}^{(n+1)} = \psi\left(\boldsymbol{\Upsilon}^{(n)}\right) - \psi(\boldsymbol{\Upsilon}^{(n)} * \boldsymbol{1})$$

$$n = n + 1$$

until termination criterion not satisfied.

## Appendix D - Additional Figures

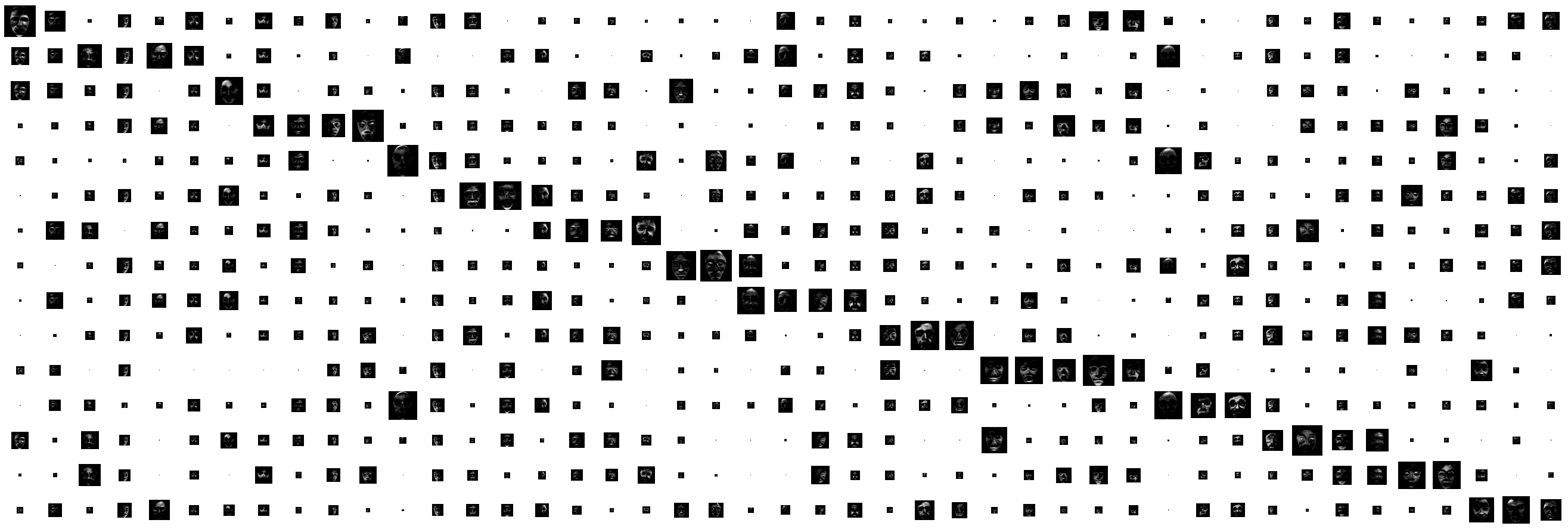

$nmf_gs;\ a^t_{vi} = 0.6,\ b^t_{vi} = 20,\ a^\lambda_{\text{small}} = 32,\ \boldsymbol{a}^\lambda_{\textbf{large}} = \mathbf{32},\ b^\lambda = 1\text{E}6$

a)

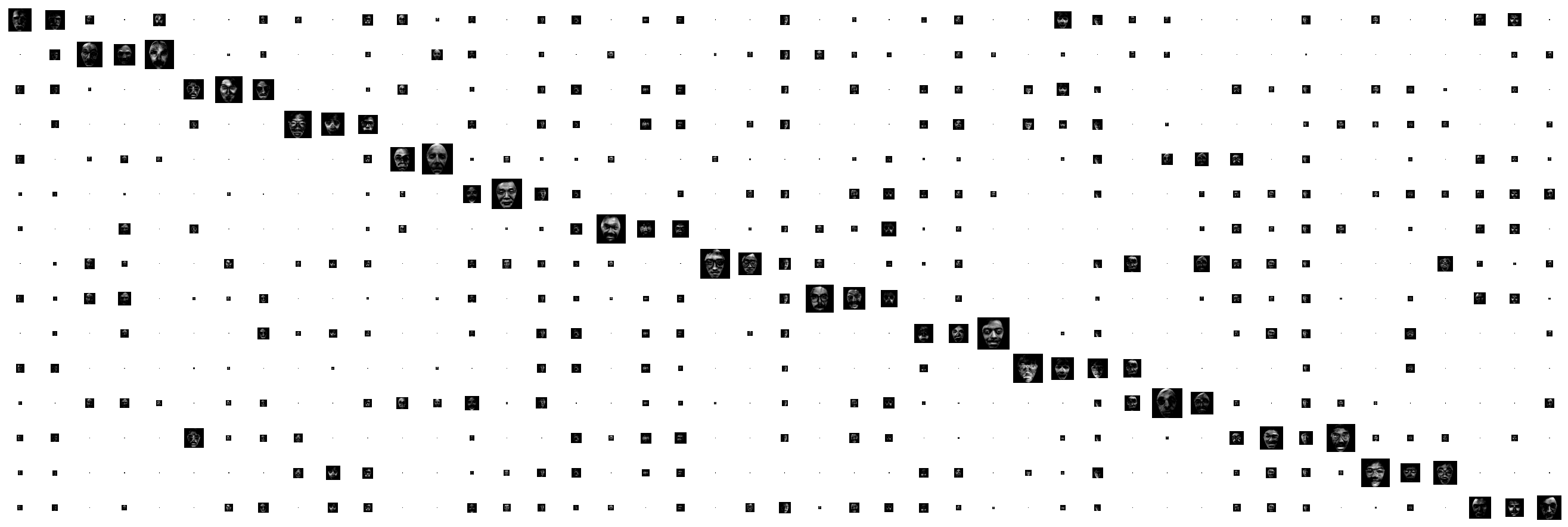

$nmf_gs;\ a^t_{vi} = 0.6,\ b^t_{vi} = 20,\ a^\lambda_{\text{small}} = 32,\ \boldsymbol{a}^\lambda_{\textbf{large}} = \mathbf{256},\ b^\lambda = 1\text{E}6$

b)

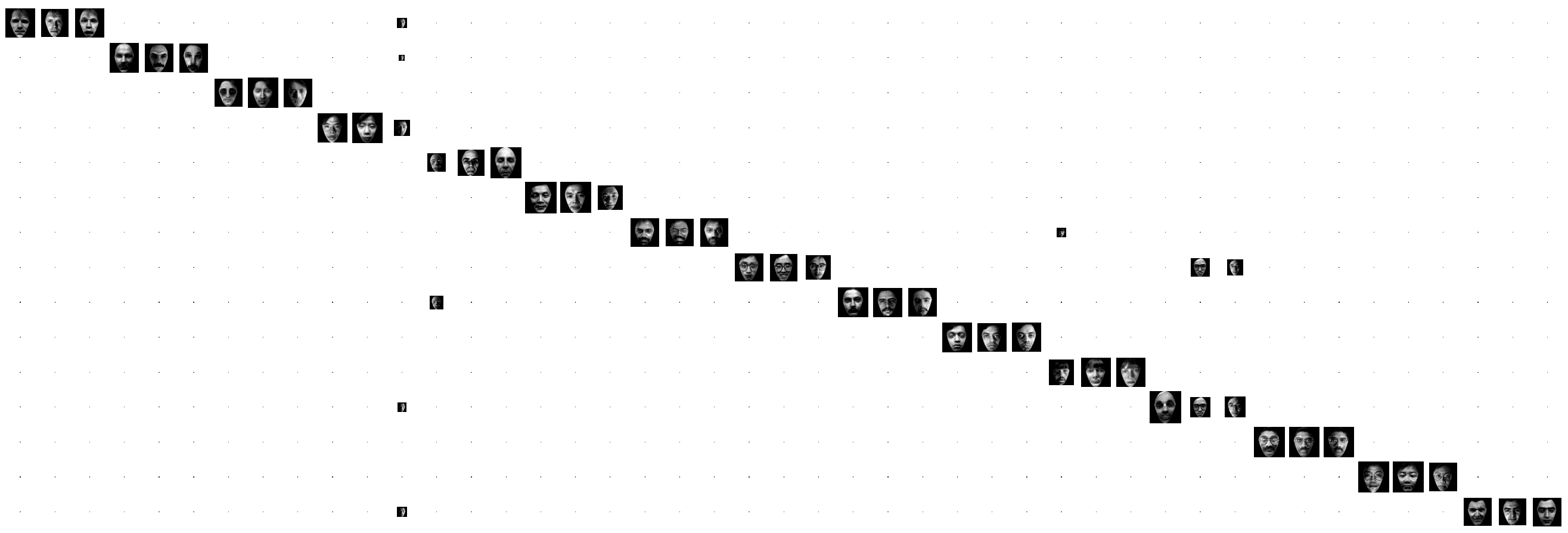

$nmf_gs;\ a^t_{vi} = 0.6,\ b^t_{vi} = 20,\ a^\lambda_{\text{small}} = 32,\ \boldsymbol{a}^\lambda_{\textbf{large}} = \mathbf{2048},\ b^\lambda = 1\text{E}6$

c)

Fig. 11. Controlling prevalences of features in labels on Yale dataset - Hinton diagrams of normalized l1 norm of the coefficient matrix $E_v$ accumulated accross labels, with corresponding features overlayed, for *nmf_gs* with increasing parameter $a^\lambda_{\text{large}}$; a) $a^\lambda_{\text{large}} = 32$, b) $a^\lambda_{\text{large}} = 256$, c) $a^\lambda_{\text{large}} = 2048$. Rows correspond to labels and columns to features.

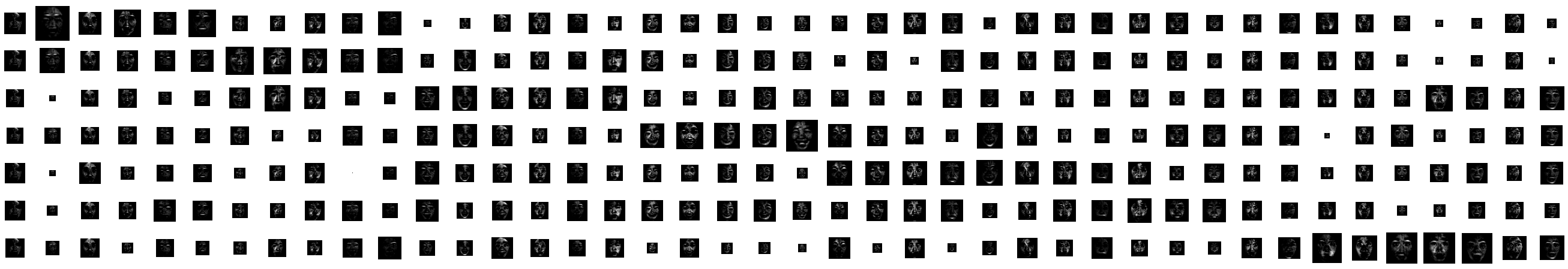

*nmf_gs*; $a_{vi}^{t} = 0.6$, $b_{vi}^{t} = 20$, $a_{\text{small}}^{\lambda} = 32$, $\boldsymbol{a}_{\textbf{large}}^{\boldsymbol{\lambda}} = \mathbf{32}$, $b^{\lambda} = 1\text{E}6$

a)

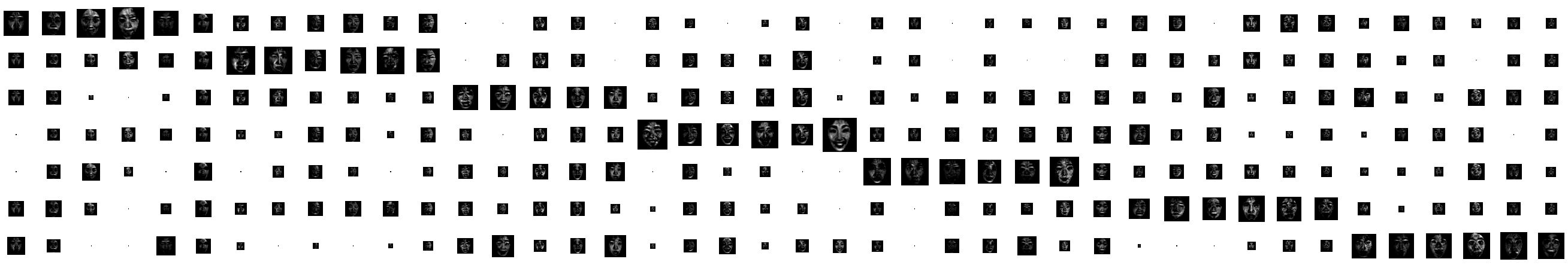

*nmf_gs*; $a_{vi}^{t} = 0.6$, $b_{vi}^{t} = 20$, $a_{\text{small}}^{\lambda} = 32$, $\boldsymbol{a}_{\textbf{large}}^{\boldsymbol{\lambda}} = \mathbf{256}$, $b^{\lambda} = 1\text{E}6$

b)

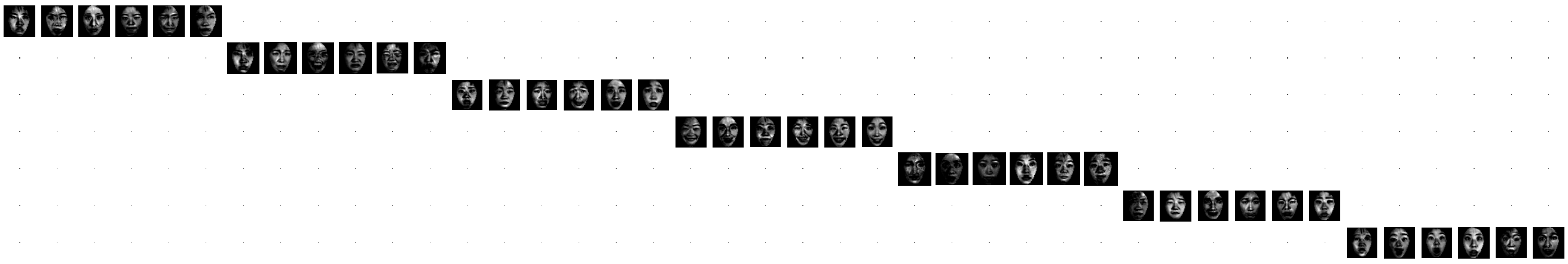

*nmf_gs*; $a_{vi}^{t} = 0.6$, $b_{vi}^{t} = 20$, $a_{\text{small}}^{\lambda} = 32$, $\boldsymbol{a}_{\textbf{large}}^{\boldsymbol{\lambda}} = \mathbf{2048}$, $b^{\lambda} = 1\text{E}6$

c)

Fig. 12. Controlling prevalences of features in labels on JAFFE dataset - Hinton diagrams of normalized l1 norm of the coefficient matrix $E_v$ accumulated accross labels, with corresponding features overlayed, for *nmf_gs* with increasing parameter $a_{\text{large}}^{\lambda}$: a) $a_{\text{large}}^{\lambda} = 32$, b) $a_{\text{large}}^{\lambda} = 256$, c) $a_{\text{large}}^{\lambda} = 2048$. Rows correspond to labels and columns to features.